\documentclass[11pt,a4paper]{article}
\usepackage[hyperref]{conll2018}
\usepackage{times}
\usepackage{url}
\usepackage{booktabs}
\usepackage{latexsym}
\usepackage{adjustbox}
\usepackage[toc,page]{appendix}
\usepackage{graphicx}
\usepackage{color}
\usepackage{multirow}
\usepackage{enumitem}
\usepackage{microtype}
\usepackage{tipa}
\usepackage{pifont}
\newcommand{\cmark}{\ding{51}}%

\newcommand{\word}[1]{\texttt{#1}}   
\newcommand{\feats}[1]{\textsf{#1}}
\newcommand{\affix}[1]{\texttt{#1}}

\usepackage{tikz}

\usepackage[small]{caption}
\aclfinalcopy

\usepackage{todonotes}

\usepackage{cleveref}

\crefname{section}{\S}{\S\S}
\Crefname{section}{\S}{\S\S}
\crefname{table}{Table}{Tables}
\crefname{figure}{Figure}{Figures}
\crefname{algorithm}{Alg.}{Algs.}
\crefname{equation}{Eq.}{Eqs.}
\crefname{appendix}{Appendix}{Appendices}
\crefformat{section}{\S#2#1#3}  

\newcommand{\ignore}[1]{}

\def\@fnsymbol#1{\ensuremath{\ifcase#1\or *\or \dagger\or \ddagger\or
   \mathsection\or \mathparagraph\or \|\or **\or \dagger\dagger
   \or \ddagger\ddagger \else\@ctrerr\fi}}

\newcommand{\nope}[0]{\textcolor{gray}{--}}

\title{The CoNLL--SIGMORPHON 2018 Shared Task: \\ Universal Morphological Reinflection}

\author{Ryan Cotterell$^{1}$ \and Christo Kirov$^1$ \and John Sylak-Glassman$^1$  \and \\ {\bf G{\'e}raldine Walther}$^2$ \and {\bf Ekaterina Vylomova$^{3}$} \and {\bf Arya D. McCarthy$^{1}$} \and \\ {\bf Katharina Kann}$^4$ \and {\bf Sabrina J. Mielke}$^1$ \and {\bf Garrett Nicolai}$^1$ \and \\ {\bf Miikka Silfverberg}$^{5,6}$ \and {\bf David Yarowsky}$^1$ \and {\bf Jason Eisner}$^1$ \and {\bf Mans Hulden}$^{5}$ \\ Johns Hopkins University$^1$ \hspace{.1cm} University of Zurich$^2$ \hspace{.1cm} University of Melbourne$^3$ \\ NYU$^4$ \hspace{.1cm} University of Colorado$^5$\hspace{.1cm} University of Helsinki$^6$\\}

\date{}

\begin{document}

\thispagestyle{plain}
\pagestyle{plain}

\maketitle
\begin{abstract}
The CoNLL--SIGMORPHON 2018 shared task on supervised learning of morphological generation featured data sets from 103 typologically diverse languages. Apart from extending the number of languages involved in earlier supervised tasks of generating inflected forms, this year the shared task also featured a new second task which asked participants to inflect words in sentential context, similar to a cloze task. This second task featured seven languages.  Task 1 received 27 submissions and task 2 received 6 submissions. Both tasks featured a low, medium, and high data condition.  Nearly all submissions featured a neural component and built on highly-ranked systems from the earlier 2017 shared task. In the inflection task (task 1), 41 of the 52 languages present in last year's inflection task showed improvement by the best systems in the low-resource setting.  The cloze task (task 2) proved to be difficult, and few submissions managed to consistently improve upon both a simple neural baseline system and a lemma-repeating baseline.

\end{abstract}

\section{Introduction}

Some of a word's syntactic and semantic properties are expressed on the word form through a process termed morphological inflection. For example, each English count noun has both singular and plural forms (\word{robot}/\word{robots}, \word{process}/\word{processes}), known as the inflected forms of the noun. Some languages display little inflection, while others possess a proliferation of forms.  A Polish verb can have nearly 100 inflected forms and an Archi verb has thousands \cite{kibrik1998archi}.

Natural language processing systems must be able to analyze and generate these inflected forms. Fortunately, inflected forms tend to be systematically related to one another. This is why English speakers can usually predict the singular form from the plural and vice versa, even for words they have never seen before: given a novel noun \word{wug}, an English speaker knows that the plural is \word{wugs}.

We conducted a competition on generating inflected forms.  This ``shared task'' consisted of two separate scenarios. In Task 1, participating systems must inflect word forms based on labeled examples. In English, an example of inflection is the conversion of a citation form\footnote{In this work we use the terms {\it citation form} and {\it lemma} interchangeably.} \word{run} to its present participle, \word{running}. The system is provided with the source form and the morphosyntactic description (MSD) of the target form, and must generate the actual target form. Task 2 is a harder version of Task 1, where the system must infer the appropriate MSD from a sentential context. This is essentially a cloze task, asking participants to provide the correct form of a lemma in context.

\section{Tasks and Evaluation}

\subsection{Task 1: Inflection}

\begin{table}
\begin{adjustbox}{width=\columnwidth}
\begin{tabular}{llll}
\toprule
{\bf Lang} & {\bf Lemma}  & {\bf Inflection} & {\bf Inflected form} \\
\midrule
\multirow{ 2}{*}{en}    & \word{hug}              & \feats{V;PST}           & \word{hugged}           \\
                             & \word{spark}            & \feats{V;V.PTCP;PRS}    & \word{sparking}         \\
\midrule
\multirow{ 2}{*}{es}    & \word{liberar}          & \feats{V;IND;FUT;2;SG}  & \word{liberar\'as}        \\
                             & \word{descomponer}      & \feats{V;NEG;IMP;2;PL}  & \word{no descompong\'ais} \\
\midrule
\multirow{ 2}{*}{de}     & \word{aufbauen}         & \feats{V;IND;PRS;2;SG}  & \word{baust auf}        \\
                             & \word{\"Arztin}	          & \feats{N;DAT;PL}         & \word{\"Arztinnen}        \\
\bottomrule
\end{tabular}
\end{adjustbox}
\caption{Example training data from task 1. Each training example maps a {\it lemma} and {\it inflection} to an {\it inflected form},
The inflection is a bundle of {\it morphosyntactic features}.  Note that inflected forms (and lemmata) can encompass multiple words. In the test data, the last column (the inflected form) must be predicted by the system.}
\label{tab:sub1data}
\end{table}

The first task was identical to sub-task 1 from the CoNLL--SIGMORPHON 2017 shared task \cite{cotterell-conll-sigmorphon2017}, but the language selection was extended from 52 languages to 103. The data sets for the overlapping languages between 2017 and 2018 were also resampled and are not identical. The task consists of morphological generation with sparse training data, something that can be practically useful for MT and other downstream tasks in NLP.  Here, participants were given examples of inflected forms as shown in \cref{tab:sub1data}.  Each test example asked participants to produce some other inflected form when given a lemma and a bundle of morphosyntactic features as input.

The training data was sparse in the sense that it included only a few inflected forms from each lemma.  That is, as in human L1 learning, the learner does not necessarily observe any complete paradigms in a language where the paradigms are large (e.g., dozens of inflected forms per lemma).\footnote{
Of course, human L1 learners do not get to observe explicit morphological feature bundles for the types that they observe.  Rather, they analyze inflected tokens in context
to discover both morphological features (including {\em inherent} features such as noun gender \cite{arnon12}) and paradigmatic structure (number of forms per lemma, number of expressed featural contrasts such as tense, number, person\ldots).}

Key points:
\begin{enumerate}
	\item The task is inflection: Given an input lemma and desired output tags, participants had to generate the correct output inflected form (a string).
	\item The supervised training data consisted of individual forms (see \cref{tab:sub1data}) that were sparsely sampled from a large number of paradigms.
        \item Forms that are empirically more frequent were more likely to appear in both training and test data (see \cref{sec:data} for details).
	\item Systems were evaluated after training on $10^2$ (low), $10^3$ (medium), and $10^4$ (high) lemma/MSD/inflected form triplets.
	\end{enumerate}

\subsection{Task 2: Inflection in Context}

The cloze test is a common exercise in an L2 instruction setting. In the cloze test, a number of words are deleted from a text and students are required to fill in the gaps with contextually plausible forms, often working from the knowledge about which lemma should be inflected. The second task of the morphology shared task presents two variations of this traditional cloze test in two tracks specifically aimed at data-driven morphology learning. 

Solving a cloze test well requires integration of many types of evidence beyond the pure capacity to inflect a word on demand. Since our training sets were gathered from actual textual resources, a good solver that accurately determines the most plausible form must implicitly combine knowledge of morphology, morphosyntax, semantics, and pragmatics. Potentially, even textual register and genre may affect the choice of correct form.  Hence, the task is both intrinsically interesting from a linguistic point of view and carries potential to support many downstream NLP applications.

\begin{figure}[!htb]
\begin{adjustbox}{width=0.9\columnwidth}
\includegraphics{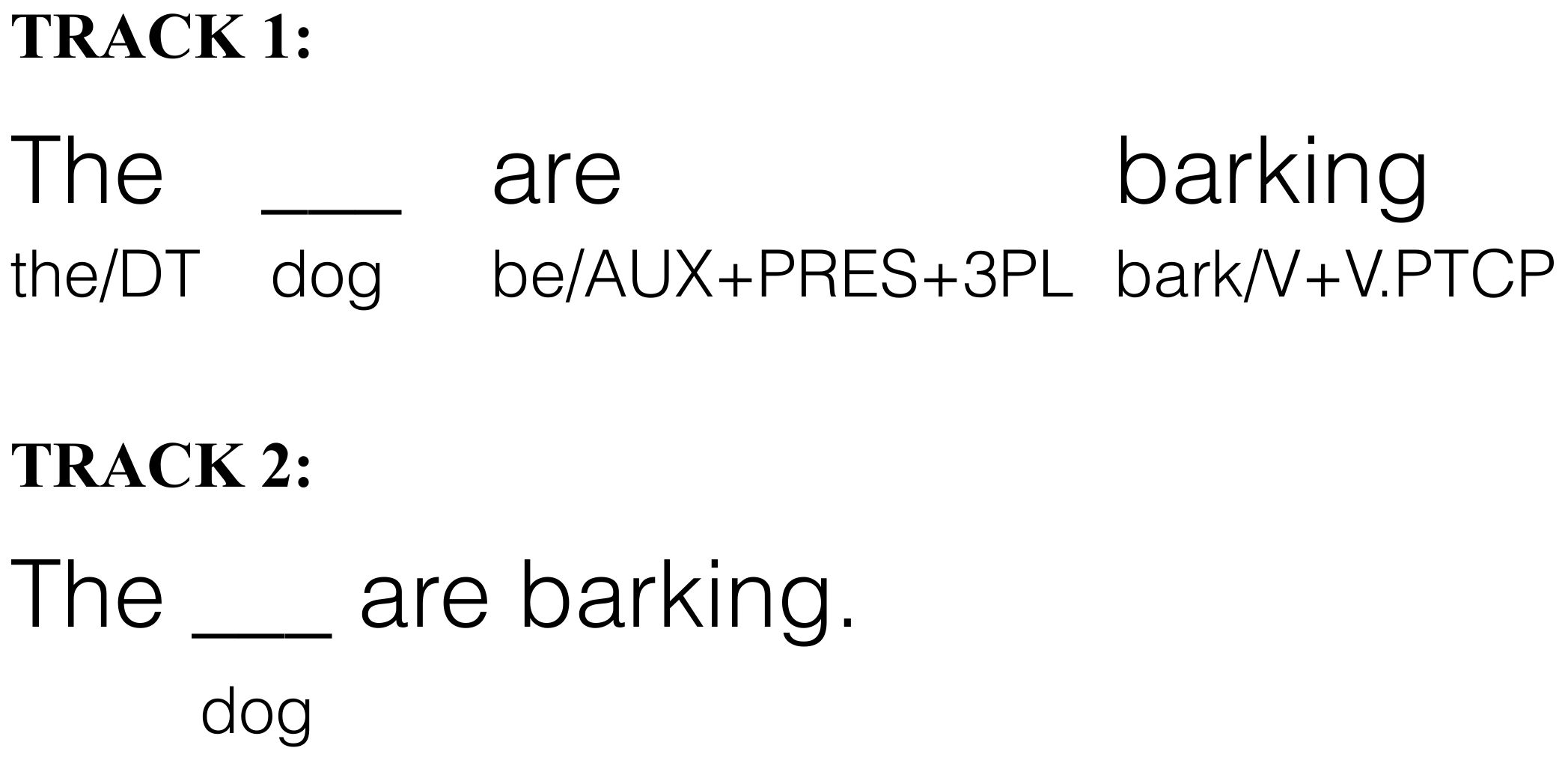}
\end{adjustbox}
\caption{Test examples for tracks 1 and 2 in the cloze task. The objective is to inflect the target lemma \word{dog} in a contextually appropriate form, which in this case is \word{dogs}. Competitors observe context word forms, their lemmata and MSDs in track 1, whereas they only observe the context word forms in track 2.}\label{fig:cloze-test}
\end{figure}

As shown in \autoref{fig:cloze-test}, both tracks supply the lemma of the omitted target word form and ask the competitors to inflect the lemma in a contextually appropriate way. In the first track, the competitors additionally see the lemmata and MSDs for all context words, whereas in the second track only the context words are available. In contrast to task~1, the MSD for the target lemma is never observed in either the first or the second track. This means that successful inflection requires the competitors to identify relevant contextual cues.

\begin{figure}[!htb]
\begin{adjustbox}{width=\columnwidth}
\includegraphics{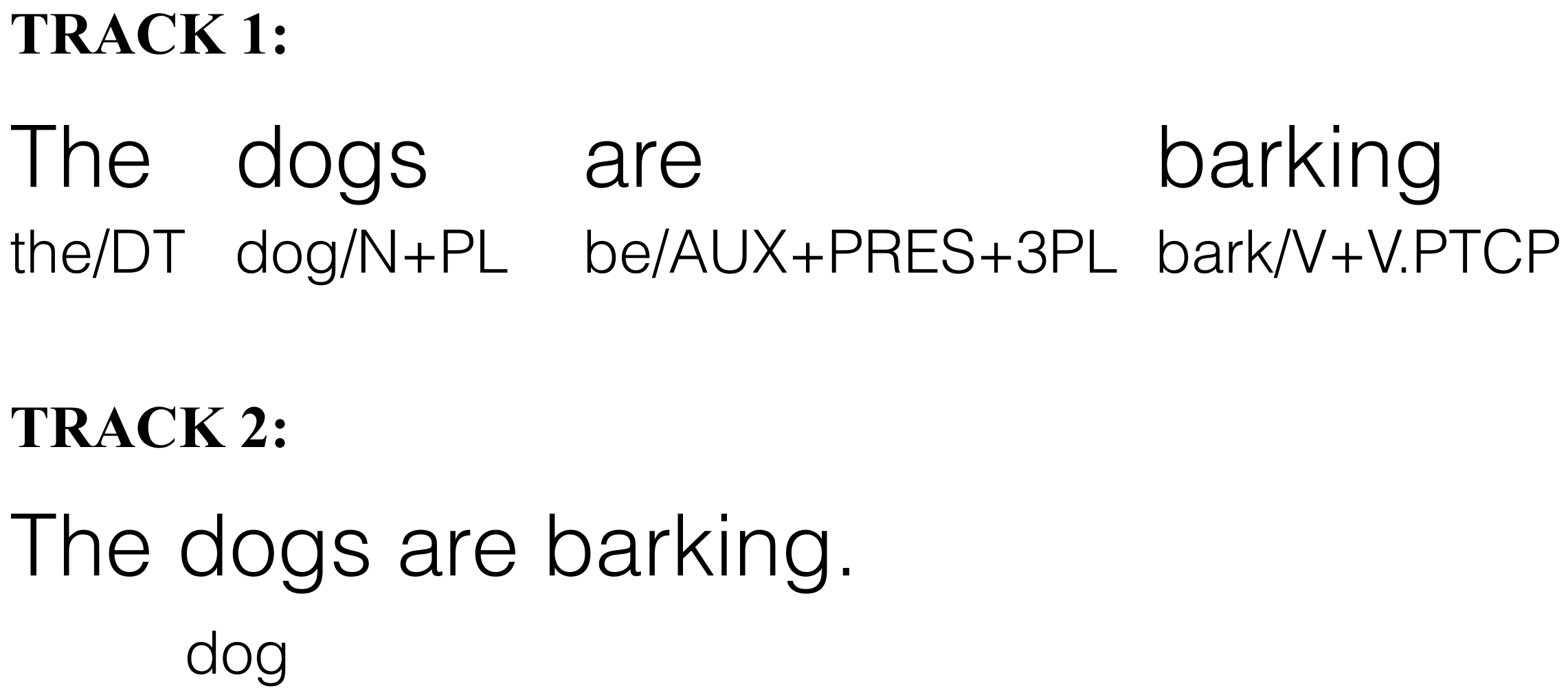}
\end{adjustbox}
\caption{Training examples for tracks 1 and 2 in the cloze task. Track 1 supplies a full morphosyntactically annotated corpus as training data, whereas track 2 only supplies lemmata for a number of selected training tokens. Remaining tokens lack annotation altogether.}\label{fig:cloze-train}
\end{figure}

As training data, the first track supplies a full morphosyntactically annotated corpus of sentences: every token is annotated with a lemma and MSD as shown in \autoref{fig:cloze-train}. In the second track, the training data identifies a number of target tokens. Lemmata are supplied for these tokens but the remaining tokens receive no MSD annotation. 

Similarly to task 1, both tracks in task 2 provide three different training data settings providing varying amounts of data: low (ca.\ $10^3$ tokens), medium (ca.\ $10^4$ tokens) and high (ca.\ $10^5$ tokens). The token counts refer to the total number of tokens in the training sets. In the first track, this allows competitors to train their systems on all available tokens. In the second track, however, only a number of tokens supply the input lemma as explained above. Thus, the effective number of training examples is smaller in the second track than in the first track. In both tracks, competitors were restricted to using only the provided training sets. For example, semi-supervised training using external data was forbidden.

Key points:
\begin{enumerate}
\item The task is inflection in context. Given an input lemma in sentential context, participants generate the correct inflected output form.
\item Two degrees of supervision are provided. In track 1, participants see context word forms and their lemmata, as well as their MSDs. In track 2, participants only witness context word forms.
\item The supervised training data, the development data, and the test data consist of sampled sentences from Universal Dependencies (UD) treebanks \cite{nivre2017universal} together with UD-provided lemmata as well as MSDs, which were converted to the UniMorph format, in track 1.
\end{enumerate}

\section{Data}\label{sec:data}

\subsection{Data for Task 1}

\paragraph{Languages}

The data for the shared task was highly multilingual, comprising 103 unique languages. Of these, 52 were shared with the 2017 shared task \cite{cotterell-conll-sigmorphon2017}. As with all but 5 of the 2017 languages (Khaling, Kurmanji Kurdish, Sorani Kurdish, Haida, and Basque), the 34 remaining 2018 languages were sourced from the English edition of
Wiktionary, a large multi-lingual crowd-sourced dictionary containing morphological paradigms for many lemmata.\footnote{\url{https://en.wiktionary.org/} (08-2016 snapshot)}

The shared task language set is genealogically diverse, including languages from $\sim$20 language stocks. Although the majority of the languages are
Indo-European, we also include two language isolates (Haida and
Basque) along with languages from Athabaskan (Navajo), Kartvelian
(Georgian), Quechua, Semitic (Arabic, Hebrew), Sino-Tibetan (Khaling),
Turkic (Turkish), and Uralic (Estonian, Finnish, Hungarian, and
Northern Sami) language families. The shared task language set is also diverse in terms
of morphological structure, with languages which use primarily
prefixes (Navajo), suffixes (Quechua and Turkish), and a mix, with
Spanish exhibiting internal vowel variations along with suffixes and
Georgian using both infixes and suffixes. The language set also
exhibits features such as templatic morphology (Arabic, Hebrew), vowel
harmony (Turkish, Finnish, Hungarian), and consonant harmony (Navajo)
which require systems to learn non-local alternations. Finally, the
resource level of the languages in the shared task set varies greatly,
from major world languages (e.g.~Arabic, English, French, Spanish,
Russian) to languages with few speakers (e.g.~Haida, Khaling). Typologically, the majority of the languages are agglutinating or fusional, with three polysynthetic languages; Haida, Greenlandic, and Navajo.\footnote{Although, some linguists \cite{baker1996polysynthesis} would exclude Navajo from the polysynthetic languages due to its lack of noun incorporation.}

\paragraph{Data Format}

For each language, the basic data consists of triples of the form
(lemma, feature bundle, inflected form), as in \cref{tab:sub1data}.
The first feature in the bundle always specifies the core part of
speech (e.g., verb).

  All features in the
bundle are coded according to the UniMorph Schema, a
cross-linguistically consistent universal morphological feature set
\cite{sylak-glassmankirov2015,sylakglassman-EtAl:2015:ACL-IJCNLP}.

\paragraph{Extraction from Wiktionary}

For each of the Wiktionary languages, Wiktionary provides a number
of tables, each of which specifies the full inflectional paradigm for
a particular lemma.  These tables were extracted using a template annotation procedure described in \cite{KIROV18.789}.

Within a language, different paradigms may have different shapes.
To prepare the shared task data, each language's
parsed tables from Wiktionary were grouped according to their tabular
structure and number of cells.  Each group represents a different type
of paradigm (e.g., verb).  We used only groups with a large number of
lemmata, relative to the number of lemmata available for the language
as a whole.
For each group, we associated a
feature bundle with each cell position in the table, by manually
replacing the prose labels describing grammatical features (e.g.~
``accusative case'') with UniMorph features (e.g.~\feats{ACC}).
This allowed us to extract triples as described in the previous section. The dataset produced by this process was sampled to create appropriately-sized data for the shared task, as described in \cref{sec:sampling}.\footnote{Full, unsampled Wiktionary parses are made available at \url{unimorph.org} on a rolling basis.}
The dataset sizes by language are given in \cref{tab:dqp1} and \cref{tab:dqp2}.

\paragraph{Sampling the Train-Dev-Test Splits.}\label{sec:sampling}

From each language's collection of paradigms, we sampled the training, development, and test sets as follows.\footnote{These datasets can be obtained from \url{https://sigmorphon.github.io/sharedtasks/2018/}} 

Our first step was to construct probability distributions over the (lemma, feature bundle, inflected form) triples in our full dataset.
For each triple, we counted how many tokens the inflected form has in the February 2017 dump of Wikipedia for that language.
To distribute the counts of an observed form over all the triples that have this token as its form, we use the syncretism resolution method of \citet{cotterell-EtAl:2018:N18-22}, training a neural network on unambiguous forms to estimate the distribution over all, even ambiguous, forms.
We then sampled 12,000 triples without replacement from this distribution.  The first 100 were taken as the low-resource training set for sub-task 1, the first 1,000 as the medium-resource training set, and the first 10,000 as the high-resource training set. Note that these training sets are nested, and that the highest-count triples tend to appear in the smaller training sets.

The final 2,000 triples were randomly shuffled and then split in half to obtain development and test sets of 1,000 forms each.  The final shuffling was performed to ensure that the development set is similar to the test set.  By contrast, the  development and test sets tend to contain lower-count triples than the training set.\footnote{This is a realistic setting, since supervised training is usually employed to generalize from frequent words that appear in annotated resources to less frequent words that do not.  Unsupervised learning methods also tend to generalize from more frequent words (which can be analyzed more easily by combining information from many contexts) to less frequent ones.}
Note that for languages that do not have enough triples for this process, we settle for omitting the higher-resource training regimes and scale down the other sizes. Details for all languages are found in \cref{tab:dqp1,tab:dqp2}.


\begin{table*}
\centering
\begin{adjustbox}{width=.9\textwidth}
\begin{tabular}{l | l | r@{\ /\ }l | r r r | r | r }
\toprule
\textbf{Language} & \textbf{Family} & \textbf{Lemmata}&\textbf{Forms} & \textbf{High} & \textbf{Medium} & \textbf{Low} & \textbf{Dev} & \textbf{Test} \\
\midrule
Adyghe & Caucasian & 1666 & 20475 & 1664/10000 & 760/1000 & 99/100 & 763/1000 & 749/1000\\
Albanian & Indo-European & 589 & 33483 & 588/10000 & 375/1000 & 84/100 & 377/1000 & 373/1000\\
Arabic & Semitic & 4134 & 140003 & 3204/10000 & 832/1000 & 99/100 & 807/1000 & 813/1000\\
Armenian & Indo-European & 7033 & 338750 & 4658/10000 & 903/1000 & 98/100 & 880/1000 & 900/1000\\
Asturian & Romance & 436 & 29797 & 432/10000 & 361/1000 & 90/100 & 368/1000 & 365/1000\\
Azeri & Iranian & 340 & 8004 & 340/6488 & 290/1000 & 79/100 & 73/100 & 81/100\\
Bashkir & Turkic & 1084 & 12168 & 1084/10000 & 662/1000 & 94/100 & 657/1000 & 651/1000\\
Basque & Isolate & 45 & 12663 & 45/10000 & 42/1000 & 24/100 & 41/1000 & 43/1000\\
Belarusian & Slavic & 1027 & 16113 & 1027/10000 & 616/1000 & 98/100 & 628/1000 & 630/1000\\
Bengali & Indo-Aryan & 136 & 4443 & 136/4243 & 134/1000 & 65/100 & 65/100 & 68/100\\
Breton & Celtic & 44 & 2294 & 44/1983 & 44/1000 & 40/100 & 38/100 & 39/100\\
Bulgarian & Slavic & 2468 & 55730 & 2133/10000 & 716/1000 & 98/100 & 742/1000 & 744/1000\\
Catalan & Romance & 1547 & 81576 & 1545/10000 & 746/1000 & 95/100 & 738/1000 & 738/1000\\
Classical-Syriac & Semitic & 160 & 3652 & 160/2396 & 160/1000 & 74/100 & 70/100 & 73/100\\
Cornish & Celtic & 9 & 469 & --- & 9/346 & 9/100 & 9/50 & 9/50\\
Crimean-Tatar & Turkic & 1230 & 7514 & 1230/7314 & 704/1000 & 94/100 & 95/100 & 95/100\\
Czech & Slavic & 5125 & 134527 & 3908/10000 & 848/1000 & 97/100 & 848/1000 & 849/1000\\
Danish & Germanic & 3193 & 25508 & 3137/10000 & 877/1000 & 100/100 & 866/1000 & 853/1000\\
Dutch & Germanic & 4993 & 55467 & 4161/10000 & 913/1000 & 100/100 & 898/1000 & 894/1000\\
English & Germanic & 22765 & 120004 & 8367/10000 & 989/1000 & 100/100 & 985/1000 & 984/1000\\
Estonian & Uralic & 886 & 38215 & 886/10000 & 587/1000 & 94/100 & 553/1000 & 577/1000\\
Faroese & Germanic & 3077 & 45474 & 2959/10000 & 857/1000 & 99/100 & 852/1000 & 865/1000\\
Finnish & Uralic & 57642 & 2490377 & 8643/10000 & 985/1000 & 100/100 & 983/1000 & 987/1000\\
French & Romance & 7535 & 367732 & 5592/10000 & 936/1000 & 98/100 & 948/1000 & 941/1000\\
Friulian & Romance & 168 & 8071 & 168/7871 & 168/1000 & 76/100 & 79/100 & 79/100\\
Galician & Romance & 486 & 36801 & 486/10000 & 421/1000 & 91/100 & 421/1000 & 423/1000\\
Georgian & Kartvelian & 3782 & 74412 & 3537/10000 & 861/1000 & 100/100 & 872/1000 & 874/1000\\
German & Germanic & 15060 & 179339 & 6797/10000 & 961/1000 & 100/100 & 945/1000 & 962/1000\\
Greek & Hellenic & 10581 & 186663 & 5130/10000 & 897/1000 & 98/100 & 915/1000 & 908/1000\\
Greenlandic & Inuit & 23 & 368 & --- & 23/268 & 23/100 & 21/50 & 21/50\\
Haida & Isolate & 41 & 7040 & 41/6840 & 41/1000 & 40/100 & 34/100 & 38/100\\
Hebrew & Semitic & 510 & 13818 & 510/10000 & 470/1000 & 95/100 & 431/1000 & 453/1000\\
Hindi & Indo-Aryan & 258 & 54438 & 258/10000 & 252/1000 & 85/100 & 254/1000 & 255/1000\\
Hungarian & Uralic & 13989 & 503042 & 7123/10000 & 963/1000 & 100/100 & 973/1000 & 978/1000\\
Icelandic & Germanic & 4775 & 76945 & 4115/10000 & 894/1000 & 100/100 & 898/1000 & 906/1000\\
Ingrian & Uralic & 50 & 1099 & --- & 50/999 & 45/100 & 30/50 & 31/50\\
Irish & Celtic & 7464 & 107298 & 5040/10000 & 906/1000 & 99/100 & 913/1000 & 893/1000\\
Italian & Romance & 10009 & 509574 & 6389/10000 & 948/1000 & 100/100 & 942/1000 & 944/1000\\
Kabardian & Caucasian & 250 & 3092 & 250/2892 & 246/1000 & 81/100 & 82/100 & 81/100\\
Kannada & Dravidian & 159 & 6402 & 159/4383 & 147/1000 & 54/100 & 53/100 & 59/100\\
Karelian & Uralic & 20 & 682 & --- & 20/582 & 20/100 & 17/50 & 18/50\\
Kashubian & Slavic & 37 & 509 & --- & 37/402 & 34/100 & 27/50 & 28/50\\
Kazakh & Turkic & 26 & 357 & --- & 26/257 & 26/100 & 22/50 & 25/50\\
Khakas & Turkic & 75 & 1200 & --- & 52/732 & 44/100 & 31/50 & 32/50\\
Khaling & Sino-TIbetan & 591 & 156097 & 584/10000 & 426/1000 & 92/100 & 411/1000 & 422/1000\\
Kurmanji & Iranian & 15083 & 216370 & 7046/10000 & 945/1000 & 100/100 & 949/1000 & 958/1000\\
Ladin & Romance & 180 & 7656 & 180/7456 & 179/1000 & 80/100 & 81/100 & 75/100\\
Latin & Romance & 17214 & 509182 & 6517/10000 & 943/1000 & 100/100 & 939/1000 & 945/1000\\
\bottomrule
\end{tabular}
\end{adjustbox}
\caption{Total number of lemmata and forms available for sampling, and number of distinct lemmata and forms present in each data condition in Task 1. Data permitting, there were 10000,1000, and 100 forms in the High, Medium, and Low conditions, respectively, and 1000 forms in each Dev and Test set.}
\label{tab:dqp1}
\end{table*}

\begin{table*}
\centering
\begin{adjustbox}{width=.9\textwidth}
\begin{tabular}{l | l | r@{\ /\ }l | r r r | r | r }
\toprule
\textbf{Language} & \textbf{Family} & \textbf{Lemmata}&\textbf{Forms} & \textbf{High} & \textbf{Medium} & \textbf{Low} & \textbf{Dev} & \textbf{Test} \\
\midrule
Latvian & Baltic & 7548 & 136998 & 5268/10000 & 930/1000 & 99/100 & 922/1000 & 923/1000\\
Lithuanian & Baltic & 1458 & 34130 & 1443/10000 & 632/1000 & 96/100 & 664/1000 & 639/1000\\
Livonian & Uralic & 203 & 3987 & 203/3787 & 203/1000 & 71/100 & 70/100 & 70/100\\
Lower-Sorbian & Slavic & 994 & 20121 & 993/10000 & 616/1000 & 96/100 & 621/1000 & 631/1000\\
Macedonian & Slavic & 10313 & 168057 & 6107/10000 & 951/1000 & 99/100 & 943/1000 & 956/1000\\
Maltese & Semitic & 112 & 3584 & 112/1560 & 112/1000 & 68/100 & 71/100 & 69/100\\
Mapudungun & Araucanian & 26 & 783 & --- & 26/602 & 26/100 & 22/50 & 23/50\\
Middle-French & Romance & 603 & 36970 & 603/10000 & 480/1000 & 92/100 & 491/1000 & 505/1000\\
Middle-High-German & Germanic & 29 & 708 & --- & 29/594 & 27/100 & 19/50 & 22/50\\
Middle-Low-German & Germanic & 52 & 1513 & --- & 52/988 & 43/100 & 30/50 & 34/50\\
Murrinhpatha & Australian & 29 & 1110 & --- & 29/973 & 28/100 & 24/50 & 24/50\\
Navajo & Athabaskan & 674 & 12354 & 674/10000 & 489/1000 & 92/100 & 491/1000 & 494/1000\\
Neapolitan & Romance & 40 & 1808 & 40/1568 & 40/1000 & 36/100 & 38/100 & 37/100\\
Norman & Romance & 5 & 280 & --- & 5/180 & 5/100 & 5/50 & 5/50\\
North-Frisian & Germanic & 51 & 3204 & 51/2256 & 51/1000 & 42/100 & 43/100 & 44/100\\
Northern-Sami & Uralic & 2103 & 62677 & 1977/10000 & 750/1000 & 97/100 & 717/1000 & 730/1000\\
Norwegian-Bokmaal & Germanic & 5527 & 19238 & 5041/10000 & 925/1000 & 100/100 & 928/1000 & 930/1000\\
Norwegian-Nynorsk & Germanic & 4689 & 16563 & 4420/10000 & 922/1000 & 99/100 & 903/1000 & 912/1000\\
Occitan & Romance & 174 & 8316 & 174/8116 & 173/1000 & 76/100 & 81/100 & 75/100\\
Old-Armenian & Indo-European & 4300 & 93085 & 3413/10000 & 837/1000 & 100/100 & 802/1000 & 822/1000\\
Old-Church-Slavonic & Slavic & 152 & 4148 & 152/2961 & 151/1000 & 78/100 & 70/100 & 76/100\\
Old-English & Germanic & 1867 & 42425 & 1795/10000 & 688/1000 & 96/100 & 708/1000 & 701/1000\\
Old-French & Romance & 1700 & 123374 & 1666/10000 & 745/1000 & 96/100 & 769/1000 & 722/1000\\
Old-Irish & Celtic & 49 & 1078 & --- & 49/851 & 38/100 & 27/50 & 26/50\\
Old-Saxon & Germanic & 863 & 22287 & 861/10000 & 514/1000 & 85/100 & 535/1000 & 494/1000\\
Pashto & Iranian & 395 & 6945 & 395/6340 & 289/1000 & 82/100 & 77/100 & 78/100\\
Persian & Iranian & 273 & 37128 & 273/10000 & 269/1000 & 82/100 & 268/1000 & 267/1000\\
Polish & Slavic & 10185 & 201024 & 5922/10000 & 935/1000 & 99/100 & 938/1000 & 942/1000\\
Portuguese & Romance & 4001 & 305961 & 3657/10000 & 905/1000 & 98/100 & 868/1000 & 865/1000\\
Quechua & Quechuan & 1006 & 180004 & 957/10000 & 515/1000 & 91/100 & 492/1000 & 506/1000\\
Romanian & Romance & 4405 & 80266 & 3351/10000 & 858/1000 & 99/100 & 854/1000 & 828/1000\\
Russian & Slavic & 28068 & 473481 & 8241/10000 & 973/1000 & 100/100 & 985/1000 & 977/1000\\
Sanskrit & Indo-Aryan & 917 & 33847 & 917/10000 & 548/1000 & 91/100 & 585/1000 & 558/1000\\
Scottish-Gaelic & Celtic & 73 & 781 & --- & 73/681 & 57/100 & 36/50 & 39/50\\
Serbo-Croatian & Slavic & 24419 & 840799 & 6726/10000 & 963/1000 & 99/100 & 965/1000 & 945/1000\\
Slovak & Slavic & 1046 & 14796 & 1046/10000 & 625/1000 & 95/100 & 590/1000 & 633/1000\\
Slovene & Slavic & 2535 & 60110 & 2368/10000 & 757/1000 & 99/100 & 760/1000 & 793/1000\\
Sorani & Iranian & 274 & 22990 & 263/10000 & 197/1000 & 74/100 & 198/1000 & 199/1000\\
Spanish & Romance & 5460 & 383390 & 4621/10000 & 906/1000 & 99/100 & 902/1000 & 922/1000\\
Swahili & Bantu & 100 & 10092 & 100/8800 & 88/1000 & 49/100 & 50/100 & 42/100\\
Swedish & Germanic & 10552 & 78407 & 6508/10000 & 952/1000 & 100/100 & 954/1000 & 970/1000\\
Tatar & Turkic & 1283 & 7832 & 1283/7632 & 736/1000 & 98/100 & 95/100 & 95/100\\
Telugu & Dravidian & 127 & 1548 & --- & --- & 18/61 & 16/50 & 16/50\\
Tibetan & Sino-Tibetan & 65 & 353 & --- & 63/158 & 56/100 & 38/50 & 38/50\\
Turkish & Turkic & 3579 & 275460 & 2876/10000 & 821/1000 & 98/100 & 849/1000 & 840/1000\\
Turkmen & Turkic & 68 & 810 & --- & 68/710 & 51/100 & 35/50 & 35/50\\
Ukrainian & Slavic & 1493 & 20904 & 1491/10000 & 722/1000 & 99/100 & 745/1000 & 736/1000\\
Urdu & Indo-Aryan & 182 & 12572 & 182/10000 & 113/1000 & 53/100 & 105/1000 & 107/1000\\
Uzbek & Turkic & 15 & 1260 & 15/1060 & 15/1000 & 15/100 & 15/100 & 15/100\\
Venetian & Romance & 368 & 18227 & 368/10000 & 339/1000 & 88/100 & 341/1000 & 340/1000\\
Votic & Uralic & 55 & 1430 & 55/1230 & 55/1000 & 50/100 & 48/100 & 47/100\\
Welsh & Celtic & 183 & 10641 & 183/10000 & 181/1000 & 78/100 & 76/100 & 80/100\\
West-Frisian & Germanic & 85 & 1429 & 85/1078 & 85/1000 & 53/100 & 62/100 & 61/100\\
Yiddish & Germanic & 803 & 7986 & 803/7356 & 581/1000 & 96/100 & 90/100 & 93/100\\
Zulu & Bantu & 566 & 39607 & 566/10000 & 450/1000 & 90/100 & 449/1000 & 443/1000\\
\bottomrule
\end{tabular}
\end{adjustbox}
\caption{Total number of lemmata and forms available for sampling, and number of distinct lemmata and forms present in each data condition in Task 1. Data permitting, there were 10000,1000, and 100 forms in the High, Medium, and Low conditions, respectively, and 1000 forms in each Dev and Test set.}
\label{tab:dqp2}
\end{table*}

\subsection{Data for Task 2}
\label{sec:data-task2}
All task 2 data sets are based on Universal Dependencies (UD) v2 treebanks \cite{nivre2017universal}. We used the data sets aimed for the 2017 CoNLL shared task on Multilingual Dependency Parsing \cite{Zeman2017} because those were available before the official UD v2 data sets.\footnote{The German 2017 CoNLL UD shared task data set is problematic: (1) there are many sentence fragments, (2) some words have complete MSDs while others are lacking MSD altogether. Therefore, we eventually decided to use the official v2 UD data sets for the German test data. These problems are not present in the official UD distribution.} For contextual inflection data sets, we retained only word forms, lemmata, part-of-speech tags and morphosyntactic feature descriptions. Dependency trees were discarded along with all other annotations present in the treebanks.

Task 2 submissions are evaluated with regard to two distinct criteria: (1) the ability of the system to reconstruct the {\it original} word form in the UD test set and (2) the ability of the system to find a {\it contextually plausible} form even if the form differs from the original one. Evaluation on plausible forms is based on manually identifying the set of contextually plausible forms for each test example. Because of the need for manual annotation, task 2 covers a more limited set of languages than task 1. In total, there are seven languages: English, Finnish, French, German, Russian, Spanish and Swedish. Token counts for the training, development and test sets are given in \autoref{tab:task2-data-size}. 

\begin{table}[!htb]
\centering{
\begin{adjustbox}{width=\columnwidth}
\begin{tabular}{lccccc}
\toprule
Language & \multicolumn{3}{c}{Train} & Dev & Test\\
         &  low  & medium &    high  &     &     \\
\midrule
English  & 1,009 & 10,016 & 100,031 & 22,509 & 22,765\\
Finnish  & 1,001 & 10,009 & 100,003 & 16,543 & 15,452 \\
French   & 1,016 & 10,004 & 100,001 & 28,304 & 14,992 \\
German   & 1,005 & 10,001 & \phantom{0}79,439  & \phantom{0}3,752 &  22,903 \\
Russian  & 1,003 & 10,020 & \phantom{0}75,964 & 11,292 &   27,935 \\
Spanish  & 1,017 & 10,035 & 100,000 & 35,209 & 27,807 \\
Swedish  & 1,007 & 10,009 & \phantom{0}66,645 & \phantom{0}7,999 & 20,808 \\
\bottomrule
\end{tabular}
\end{adjustbox}
}
\caption{Token counts of the training, development and test sets for task 2.}\label{tab:task2-data-size}
\end{table}

\begin{table}[!htb]
\centering{
\begin{tabular}{lccccc}
\toprule
Language & Dev & Test\\
\midrule
English  & 2,489 & 993\\
Finnish  & 1,881 & 787 \\
French   & 1,655 & 491 \\
German   & \phantom{0}333 & 989 \\ 
Russian   & 1,181 & 996 \\
Spanish  & 2,268 & 713 \\
Swedish  & \phantom{0}573 & 940 \\
\bottomrule
\end{tabular}
}
\caption{Counts of target lemmata to be inflected in the development and test sets for task 2.}
\end{table}

\paragraph{Data Conversion}
Some of the UD treebanks required slight modifications in order to be suitable for reinflection.   
In the Finnish data sets, lemmata for compound words included morpheme boundaries, for example \word{muisti\#kapasiteetti} `memory capacity'. The morpheme boundary symbols were deleted. In the Russian treebanks, all lemmata were written completely in upper case letters. These were converted to lower case.\footnote{We used the Python3 function \texttt{string.lower}.}

\paragraph{Manual annotation}

To produce the complete list of ``plausible forms'' annotators were given complete UniMorph inflection tables for the center lemma for each sentence and were asked to check off all forms that are ``grammatically plausible'' in the particular context.  For example, given an original sentence {\bf We saw the dog}, the form {\bf dogs} would be contextually plausible and would be annotated into the test set. For pro-drop languages and short sentences, it is sometimes the case that all or most indicative, conditional, and future forms of a verb are acceptable when the subject is omitted and agreement is unknown. For example, consider the Spanish sentence from the test data:

\bigskip

\begin{adjustbox}{width=\columnwidth}
\noindent\begin{tabular}{lllll}
$\underline{\phantom{\textrm{Será}}}$ & la & mejor & de  & Primera \\
{\bf ser} \\
`{\bf to be}' & `the' & `best' & `of' & `premier (league)' \\
\end{tabular}
\end{adjustbox}

\bigskip

Obviously, almost any person, tense, and aspect of the verb `to be' will be appropriate for this limited context (\word{ser\'ia} `I would be', \word{fue} `he/she/it was', \word{eres} `you are', \ldots). Of course, depending on the genre of the text, some would be highly implausible, but the annotation intends to capture morphosyntactic rather than semantic and pragmatic felicity.

We had one annotator for each test set, with the exception of French, in which, due to practical difficulties in finding a native speaker annotator, we did not annotate the plausible forms and instead used the original sentences.

When forming the final test sets, all test examples with more than $5$ contextually plausible word form alternatives were filtered out. This was done because a large number of plausible word forms was deemed to raise the risk of annotation errors. A threshold of $5$ plausible forms was chosen because it means that all languages have test sets greater than $700$ examples. The test set for French is smaller but this is not due to manual annotations.

\begin{figure}[!htb]
\begin{adjustbox}{width=0.85\columnwidth}
\includegraphics[angle=90]{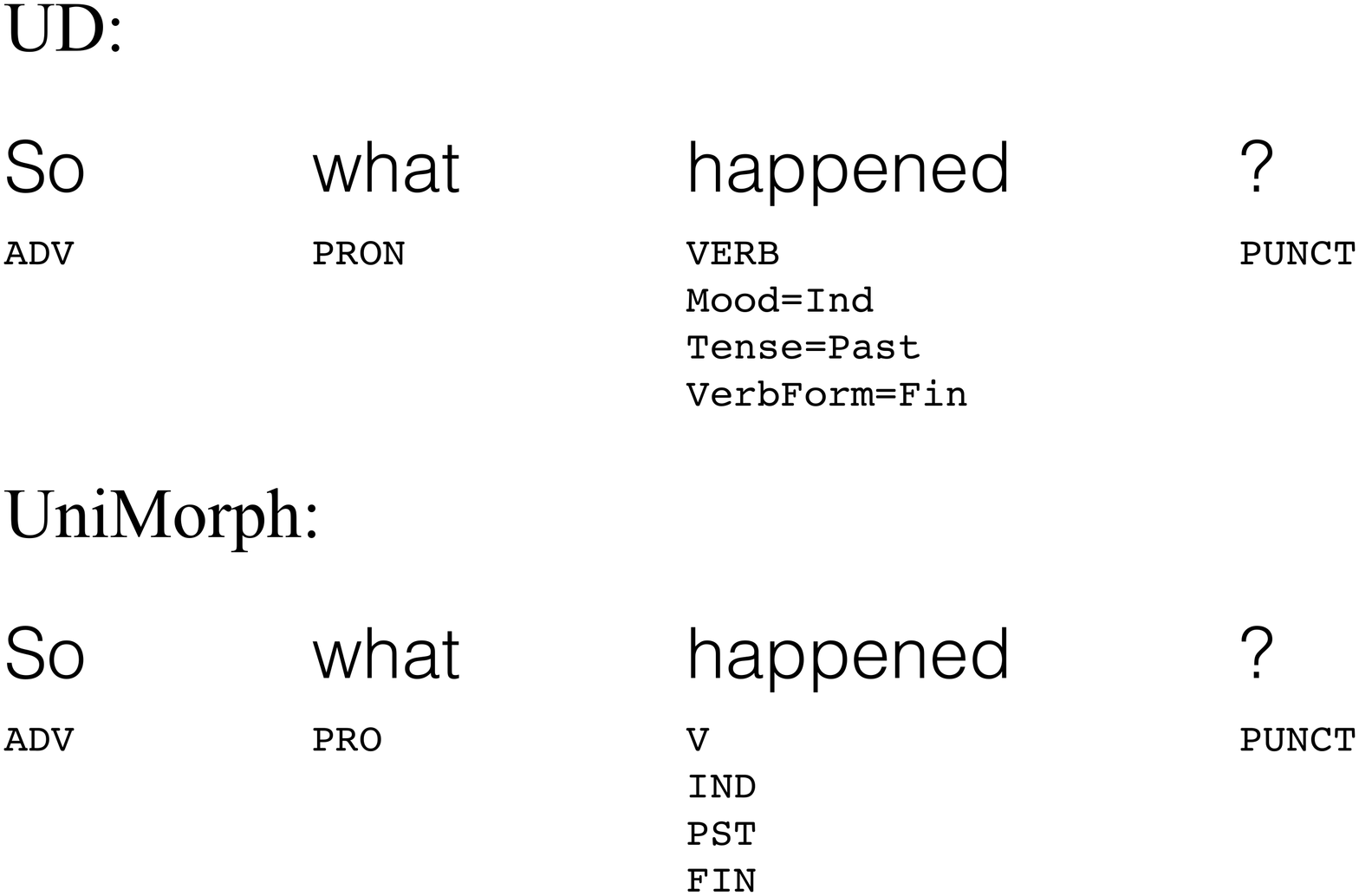}
\end{adjustbox}
\caption{A morphosyntactically annotated sentence from the original UD treebank for English and the result of an automatic conversion into the UniMorph annotation schema.}\label{fig:ud2unimorph}
\end{figure}

\paragraph{Sampling examples} The data sets for each language are based on UD treebanks for the given language. We preserved UD splits into training, development and test data. 

For each UD treebank, we first formed sets of training, development and test candidate sentences. A sentence was a candidate for the shared task data set if it contained a token found in the UniMorph resource for the relevant language; or more precisely, a token whose word form, lemma and MSD occur in a same UniMorph inflection table.

We limited target tokens to tokens present in the UniMorph resource in order to facilitate manual annotation of data sets. In particular, we limited the set of possible target MSDs to MSDs which occur in the Unimorph resource. This was necessary to avoid a prohibitively large number of contextually plausible inflections in certain languages. For example, Finnish includes a number of clitics (\affix{ko/k\"a}, \affix{kin}, \affix{han/h\"an}, \affix{pa/p\"a}, \affix{s}, \affix{kaan/k\"a\"an}) which can be appended relatively freely to word forms. Combinations of clitics are also possible. This easily leads to hundreds of word forms which can be contextually plausible. Restricting the MSDs of a possible output form to the more limited set of MSDs occurring in the UniMorph resource made the selection of plausible forms far more manageable from an annotation perspective. 

Training data sets were formed from candidate sentences simply by sampling a suitable number of sentences from the candidate sets in order to achieve the desired token counts $10^3$, $10^4$, and $10^5$ for the low, medium, and high data settings, respectively. For German and Russian, all candidate sentences were used in the high data setting, although this was not sufficient to create a training set of $10^4$ tokens. The training sets for German and Russian are, therefore, smaller than those for the other languages. For the development sets, we used all available candidate sentences for all of the languages. 

For the test data, we first formed a set of candidate sentences so that the combined number of target tokens in the test sets was 1,000.\footnote{For French, there were only 491 target tokens in the entire UD test data set. Those were used as the test data.} Target tokens in these initial test sets were then manually annotated with additional contextually plausible word forms. 

\paragraph{MSD conversion} Sampling of training, development and test examples was based on comparing UD word forms, lemmata and MSDs to equivalents in UniMorph paradigms. Therefore, it was necessary to convert the morphosyntactic annotation in the UD data sets into UniMorph morphosyntactic annotation. We used deterministic tag conversion rules to accomplish this. An example of a source UD sentence and a target UniMorph sentence is shown in \autoref{fig:ud2unimorph}. 

Since the selection of languages in task 2 is small and we do not attempt to correct annotation errors in the UD source materials, conversion between UD and UniMorph morphosyntactic descriptions is generally straightforward.\footnote{\newcite{McCarthy2018} present more principled and far more complete work on conversion between the UD and UniMorph resources for the full range of languages at the intersection of UD and UniMorph resources.} However, UD descriptions are more fine-grained than their UniMorph equivalents. For example, UD denotes lexical features such as noun gender which are inherent features of a lexeme possessed by all of its word forms. Such inherent features are missing from UniMorph which exclusively annotates inflectional morphology \cite{McCarthy2018}. Therefore, UD features which lack correspondents in the UniMorph tagging schema were simply dropped during conversion.

\section{Baselines}

\subsection{Task 1 Baseline}

The baseline system provided for task 1 was based on the observation that, for a large number of languages, producing an inflected form from an input citation form can often be done by memorizing the suffix changes that occur in doing so, assuming enough examples are seen \cite{liu2016}.  For example, in witnessing a Finnish inflection of the noun \word{koti} `home' in the singular elative case as \word{kodista}, a number of transformation rules can be extracted that may apply to previously unseen nouns:

\bigskip

\begin{verbatim}
    $koti$
    $kodista$  N;IN+ABL;SG
\end{verbatim}

\bigskip

In this example, the following transformation rules are extracted:

\bigskip

\begin{center}
\small
\begin{tabular}{ll}
\affix{\$} $\rightarrow$ \affix{sta\$} & \affix{i\$} $\rightarrow$ \affix{ista\$} \\
\affix{ti\$} $\rightarrow$ \affix{dista\$} & \affix{oti\$} $\rightarrow$ \affix{odista\$} \\
\affix{koti\$} $\rightarrow$ \affix{kodista\$} \\
\end{tabular}
\normalsize
\end{center}

\bigskip

Such rules are then extracted from each example inflection in the training data.  At generation time, the longest matching left hand side of a rule is identified and applied to the citation form.  For example, if the Finnish noun \word{luoti} `bullet' were to be inflected in the elative (\feats{N;IN+ABL;SG}) using only the extracted rules given above, the transformation \affix{oti\$} $\rightarrow$ \affix{odista\$} would be triggered, producing the output \word{luodista}. In case there are multiple candidate rules of equally long left hand sides that all match, ties are broken by frequency---i.e.\ the rule that has been witnessed most times in the training data applies.

Since languages may also use prefixing as a inflectional strategy, a similar process is applied to any identified prefix changes.  Identifying which parts of a change in a word form correspond to a prefix and which are considered suffixes requires alignment of the citation form and the output form, which is performed as a preliminary step. We refer the reader to \newcite{cotterell-conll-sigmorphon2017} for a detailed description of the baseline system.

\subsection{Task 2 Baseline} 

\paragraph{Neural Baseline} The neural baseline system is an encoder-decoder reinflection system with attention inspired by \newcite{kann-schutze:2016:P16-2}. The crucial difference is that the reinflection is conditioned on sentence context. This is accomplished by conditioning the encoder on embeddings of context words in track 2 and context words, their lemmata and their MSDs in track 1. 

\begin{figure}[!htb]
\centering{
\begin{adjustbox}{width=0.75\columnwidth}
\includegraphics{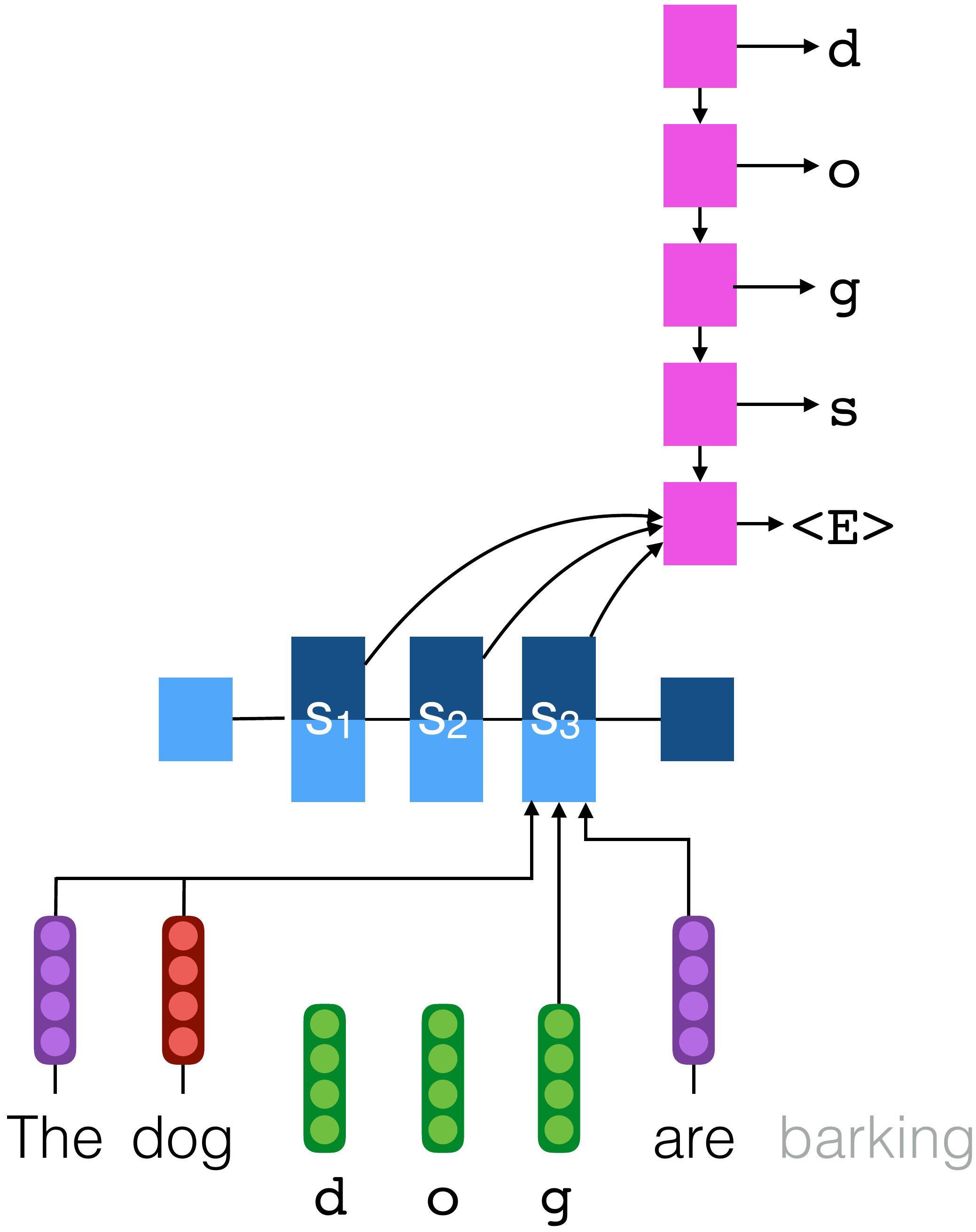}
\end{adjustbox}}
\caption{The neural baseline system for track 2 of task 2: A bidirectional LSTM encoder, conditioned on embeddings of the left context word {\bf The}, right context word {\bf are} and a whole token embedding of the lemma {\bf dog}, is used to encode the character sequence ({\bf d}, {\bf o}, {\bf g}) into representation vectors $s_1$, $s_2$ and $s_3$. An LSTM decoder with an attention mechanism generates the contextually appropriate output word form {\bf dogs}. The neural baseline system for track 1 is very similar but the encoder is conditioned on embeddings of the context words, context lemmata and context MSDs.}
\end{figure}

The neural baseline system takes as input 
\begin{enumerate}
\item A lemma $l = l_1,...,l_m$,
\item a left and right context word form $w_{{\rm L}}$ and $w_{{\rm R}}$, respectively.
\item a left and right context lemma $l_{{\rm L}}$ and $l_{{\rm R}}$, respectively (only in track 1) and
\item a left and right context MSD $m_{{\rm L}}$ and $m_{{\rm R}}$, respectively (only in track 1).
\end{enumerate}
The neural baseline system produces an inflected form $w = w_1,...,w_n$ of the lemma  as output.

The input characters $l_i$ are first embedded: $l_i \mapsto {\rm E}(l_i)$. Then, context words ($w_{\rm L}$ and $w_{\rm R}$) for both tracks, as well as context lemmata ($l_{\rm L}$ and $l_{\rm R}$) and MSDs ($m_{\rm L}$ and $m_{\rm R}$) for track 1 are also embedded: $w_X \mapsto {\rm E}(w_X)$, $l_X \mapsto {\rm E}(l_X)$ and $m_X \mapsto {\rm E}(m_X)$. The system also a uses the whole token embedding of the input lemma $l$: $l \mapsto {\rm E}(l)$.

A bidirectional LSTM encoder is used to encode the lemma into representation vectors. In order to condition the encoder on the sentence context of the lemma, the encoder input vector $e_i$ for character $l_i$ is 
\begin{enumerate}
\item a concatenation of embeddings for the context word forms, context lemmata, context MSDs, input lemma and input character: $e_i = [{\rm E}(w_{\rm L});{\rm E}(l_{\rm L}); {\rm E}(m_{\rm L});{\rm E}(l);$ ${\rm E}(w_{\rm R});{\rm E}(l_{\rm R}); {\rm E}(m_{\rm R});{\rm E}(l_i)]$ for track 1, and
\item a concatenation of embeddings for the context word forms, input lemma and input character: $e_i = [{\rm E}(w_{\rm L});{\rm E}(l);{\rm E}(w_{\rm R});{\rm E}(l_i)]$ for track 2.
\end{enumerate}

The input vectors $e_1,...,e_m$ are then encoded into representations $s_1,...,s_m$ by a bidirectional LSTM encoder. Finally, a decoder with additive attention \cite{vaswani2017attention} is used for generating the output word form $w = w_1,...,w_n$ based on the representations $s_1,...,s_m$.

The baseline system uses 100-dimensional embeddings and the LSTM hidden dimension for both the encoder and decoder is of size 100. Both encoder and decoder LSTM networks are single layer networks. The additive attention network is a 2-layer feed-forward network with hidden dimension 100 and $\tanh$ nonlinearity.  

The baseline system is trained for 20 epochs in both tracks and under all data settings using Adam \cite{DBLP:journals/corr/KingmaB14}. During training, 30\% dropout is applied on all input and recurrent connections in the encoder and decoder LSTM networks. Whole token embeddings for the input lemma, context word forms, lemmata and MSDs are dropped with a probability of 10\%.

\paragraph{Copy Baseline} The second baseline is very straightforward. It simply copies the input lemma into the output. The system is based on the observation that in many languages the lemma form is quite common. In some languages, such as English, this baseline is in fact quite difficult to beat when the  training set is small.

\begin{table*}[!htb]
\begin{adjustbox}{width=\textwidth}
\begin{tabular}{lcccccc}
\toprule
            & predict MSD & subword context & context RNN & context attention & multilingual & beam search \\
            \midrule
 BME-HAS \cite{acs:2018:K18-20}                                    & \nope  & \cmark & \cmark & \nope  & \nope  & \nope  \\
 COPENHAGEN  \cite{kementchedjhieva-bjerva-augenstein:2018:K18-20} & \cmark & \nope  & \cmark & \nope  & \cmark & \nope  \\
 CUBoulder \cite{liu-EtAl:2018:K18-20}                             & \cmark & \nope  & \nope  & \nope  & \nope  & \nope  \\
 NYU \cite{kann-lauly-cho:2018:K18-20}                             & \nope  & \cmark & \cmark & \cmark & \nope  & \nope  \\
 UZH  \cite{makarov-clematide:2018:K18-20}                         & \nope  & \cmark & \nope  & \nope  & \nope  & \cmark \\
 \bottomrule
 \end{tabular}
 \end{adjustbox}
  \caption{Features of Task 2 systems.}\label{tab:task2-design}
 \end{table*}

\section{Results}

The CoNLL--SIGMORPHON 2018 shared task received submissions from 15
teams with members from 17 universities or institutes (\cref{tab:teams}). Many of the teams submitted more than one system, yielding a total of 33 unique systems entered---27 for task 1, and 6 for task 2. In addition, baseline systems provided by the organizers for both tasks were also evaluated.

\begin{table*}
  \begin{adjustbox}{width=2\columnwidth}
  \begin{tabular}{l l l } \toprule
    Team & Institute(s) & System Description Paper  \\ \midrule
AXSEMANTICS$^1$    & AX Semantics                                  & \newcite{madsack-EtAl:2018:K18-20} \\
BME$^1$/BME-HAS$^2$ & Budapest University of Technology and Economics / Hungarian Academy of Sciences & \newcite{acs:2018:K18-20} \\
COPENHAGEN$^2$     & University of Copenhagen                      & \newcite{kementchedjhieva-bjerva-augenstein:2018:K18-20} \\ 
CUBoulder$^2$      & University of Colorado, Boulder               & \newcite{liu-EtAl:2018:K18-20} \\
HAMBURG$^1$        & Universit\"at Hamburg                         & \newcite{schroder-EtAl:2018:K18-20} \\
IITBHU$^1$         & IIT (BHU) Varanasi / IIIT Hyderabad           & \newcite{sharma-katrapati-sharma:2018:K18-20} \\
IIT-VARANASI$^1$   & Indian Institute of Technology (BHU) Varanasi & \newcite{jain-singh:2018:K18-20} \\
KUCST$^1$          & University of Copenhagen, Centre for Language Technology & \newcite{agirrezabal:2018:K18-20} \\
MSU$^1$            & Moscow State University                       & \newcite{sorokin:2018:K18-20} \\
NYU$^2$            & New York University                           & \newcite{kann-lauly-cho:2018:K18-20} \\
RACAI$^1$          & Romanian Academy                              & \newcite{dumitrescu-borocbs:2018:K18-20} \\
TUEBINGEN-OSLO$^1$ & University of Oslo / University of T\"ubingen & \newcite{rama-ccoltekin:2018:K18-20} \\
UA$^1$             & University of Alberta                         & \newcite{najafi-EtAl:2018:K18-20} \\
UZH$^{1,2}$        & University of Zurich                          & \newcite{makarov-clematide:2018:K18-20} \\
WASEDA$^1$         & Waseda University                             & \newcite{fam-lepage:2018:K18-20} \\
\bottomrule
  \end{tabular}
  \end{adjustbox}
  \caption{Participating teams, member institutes, and the corresponding system description papers. In the results and the main text, team submissions have an additional integer index to distinguish between multiple submissions by one team. The numbers at each abbreviated team name show whether teams participated in task 1, task 2, or both.}
  \label{tab:teams}
\end{table*}

\subsection{Task 1 Results}

\begin{table}[t]
\newcommand{\incomplete}[1]{\textit{\color{gray}#1}}
\begin{adjustbox}{width=.95\columnwidth}
\begin{tabular}{lccc}
\toprule
& \multicolumn{1}{c}{High} & \multicolumn{1}{c}{Medium} & \multicolumn{1}{c}{Low} \\
\midrule
uzh-01 & \textbf{96.00 / 0.08} & \textbf{86.64 / 0.26} & 57.18 / 1.00\\
uzh-02 & 95.97 / 0.08 & 86.38 / 0.27 & \textbf{57.21 / 1.02}\\
bme-02 & 94.66 / 0.11 & 67.26 / 0.88 & 2.43 / 6.91\\
iitbhu-iiith-01 & 94.43 / 0.11 & 82.90 / 0.34 & 49.79 / 1.18\\
iitbhu-iiith-02 & 94.43 / 0.11 & 84.19 / 0.32 & 52.60 / 1.10\\
bme-03 & 93.97 / 0.12 & 67.36 / 0.75 & 3.63 / 6.75\\
bme-01 & 93.88 / 0.12 & 67.43 / 0.75 & 3.74 / 6.72\\
msu-04 & 91.87 / 0.23 & 76.40 / 0.55 & 31.40 / 2.16\\
iit-varanasi-01 & 91.73 / 0.16 & 70.17 / 0.66 & 23.33 / 2.40\\
waseda-01 & 91.12 / 0.19 & 77.38 / 0.67 & 44.09 / 1.68\\
msu-03 & 90.52 / 0.25 & 75.74 / 0.55 & 25.86 / 2.38\\
axsemantics-01 & 84.19 / 0.40 & \incomplete{58.00 / 1.10} & \incomplete{72.00 / 0.96} \\
msu-02 & 82.68 / 0.41 & 69.45 / 0.79 & 41.61 / 1.86\\
racai-01 & \incomplete{79.93 / 0.43} & \textcolor{gray}{\ \ --- / ---} & \textcolor{gray}{\ \ --- / ---}\\
hamburg-01 & 77.53 / 0.44 & 74.03 / 0.54 & 40.28 / 1.45\\
axsemantics-02 & 74.77 / 0.68 & 60.00 / 1.03 & \incomplete{14.89 / 3.89}\\
msu-01 & 74.33 / 0.78 & \incomplete{64.57 / 0.93} & \textcolor{gray}{\ \ --- / ---}\\
tuebingen-oslo-03 & 63.05 / 1.15 & 30.98 / 2.25 & 1.39 / 5.70\\
tuebingen-oslo-02 & 56.60 / 1.34 & 29.72 / 2.36 & 4.43 / 5.06\\
kucst-01 & \incomplete{54.37 / 1.57} & \incomplete{32.28 / 2.23} & \incomplete{2.79 / 5.28}\\
tuebingen-oslo-01 & 49.52 / 1.67 & 20.97 / 2.81 & 0.00 / 7.94\\
ua-08 & \textcolor{gray}{\ \ --- / ---} & \textcolor{gray}{\ \ --- / ---} & 53.22 / 1.35\\
ua-05 & \textcolor{gray}{\ \ --- / ---} & \textcolor{gray}{\ \ --- / ---} & 50.53 / 1.34\\
ua-06 & \textcolor{gray}{\ \ --- / ---} & \textcolor{gray}{\ \ --- / ---} & 49.73 / 1.46\\
ua-03 & \textcolor{gray}{\ \ --- / ---} & \textcolor{gray}{\ \ --- / ---} & 44.82 / 1.45\\
ua-02 & \textcolor{gray}{\ \ --- / ---} & \textcolor{gray}{\ \ --- / ---} & \incomplete{41.61 / 2.47}\\
ua-07 & \textcolor{gray}{\ \ --- / ---} & \textcolor{gray}{\ \ --- / ---} & \incomplete{39.52 / 1.76}\\
ua-01 & \textcolor{gray}{\ \ --- / ---} & \textcolor{gray}{\ \ --- / ---} & 38.22 / 2.02\\
ua-04 & \textcolor{gray}{\ \ --- / ---} & \textcolor{gray}{\ \ --- / ---} & 21.25 / 3.43\\
\midrule
baseline & 77.42 / 0.51 & 63.53 / 0.90 & 38.89 / 1.88\\
\midrule
oracle-fc & 99.87 / \ \ \ \nope\ \ \ & 98.27 / \ \ \ \nope\ \ \ & 77.23 / \ \ \ \nope\ \ \ \\
oracle-e & 98.90 / \ \ \ \nope\ \ \ & 93.74 / \ \ \ \nope\ \ \ & 74.88 / \ \ \ \nope\ \ \ \\
\bottomrule
\end{tabular}
\end{adjustbox}
\setlength\belowcaptionskip{-5pt}
\caption{Task 1 results: Per-form accuracy (in percentage points) and average Levenshtein distance from the correct form (in characters), averaged across the 103 languages with all languages weighted equally. The columns represent the different training size conditions. Rows are sorted by accuracy under the ``High'' condition. Numbers in bold are the best accuracy in their category. Greyed-out cells represent partial submissions that did not provide output for every language, and thus do not have comparable mean scores. The per-language performance of these systems can be found in the Appendix.}\label{tab:t1perfsum}
\end{table} 

The relative system performance is described in \cref{tab:t1perfsum}, which show the average per-language accuracy of each system by resource condition. The table reflects the fact that some
teams submitted more than one system (e.g.~UZH-1 \& UZH-2 in the table). Learning curves for each language across conditions are shown in \cref{tab:bestbylanguagep1,tab:bestbylanguagep2}, which indicates the best per-form accuracy achieved by a submitted system. Full results can be found in \cref{appends}. Newer approaches led to better overall results in 2018 compared to 2017. In the low-resource condition, 41 (80\%) of the 52  languages shared across years saw improvement in top system performance.

In the lower data conditions, encoder-decoder models are known to perform worse than the baseline model due to data sparsity.  One way to work around this weakness is to learn sequences of edit operations instead of a standard string-to-string transduction, a strategy which was used by teams last year and this year (AX SEMANTICS, UZH, HAMBURG, MSU, RACAI).  Another strategy is to create artificial training data that biases the neural model toward copying \cite{kann-schutze:2017:K17-20,bergmanis-EtAl:2017:K17-20,silfverberg-EtAl:2017:K17-20,zhou-neubig:2017:K17-20,nicolai-EtAl:2017:K17-20}, which was also employed this year (TUEBINGEN-OSLO, WASEDA).  Learning edit sequences requires input/output alignment, often as a preliminary step. The UZH submissions, which attained the highest average accuracy on the higher data conditions, built upon ideas in their last year's submission \cite{makarov-ruzsics-clematide:2017:K17-20}, which had used such a separate alignment step followed by the application of an edit sequence. Their 2018 submission included edit distance alignment as part of the training loss function in the model, producing an end-to-end model. Another alternative to the edit sequence model is to use pointer generator networks, introduced by \cite{see2017get} for text summarization, which also allow for copying parts of the input. This was employed by IITBHU. BME used a modified attention model that attended to both the lemma sequence and the tag sequence, which worked well in the high data condition, but, being without models of data augmentation or edit sequences, it suffered in the low data setting. In general, systems that included edit sequence generation or data augmentation fared significantly better in the low data settings. The HAMBURG submission attempted to learn similarities between characters based on rendering them visually using a font, with the intent to discover similarities such as those between {\bf a} and {\bf \"a}, where the former is usually a low back vowel, and the latter a fronted version. Ensembling was also a popular choice to improve system performance. The UA system combined multiple models, both neural and non-neural, and focused on performance in the low data setting.

Even though the top-ranked systems used some form of ensembling to improve performance, different teams relied on different overall approaches. As a result, submissions may contain some amount of complementary information, so
that a global ensemble may improve
accuracy. As in 2017, we present an upper bound on the possible performance of
such an ensemble. \cref{tab:t1perfsum} 
includes an ``Ensemble Oracle'' system (oracle-e) that gives the correct
answer if \emph{any} of the submitted systems is correct. The oracle
performs significantly better than any one system in both the Medium
($\sim$10\%) and Low ($\sim$25\%) conditions. This suggests that the
different strategies used by teams to ``bias'' their systems in an effort
to make up for sparse data lead to substantially different
generalization patterns.

As in 2017, we also present a second ``Feature Combination''
Oracle (oracle-fc) that gives the correct answer for a given test
triple iff its feature bundle appeared in training (with any lemma).
Thus, oracle-fc provides an upper bound on the performance of systems
that treat a feature bundle such as \feats{V;SBJV;FUT;3;PL} as atomic.
In the low-data condition, this upper bound was 77\%, meaning
that 23\% of the test bundles had never been seen in training data.
Nonetheless, systems should be able to make some accurate predictions
on this 23\% by decomposing each test bundle into individual
morphological features such as \feats{FUT} (future) and \feats{PL}
(plural), and generalizing from training examples that involve
those features.
  For example, a particular feature or
sub-bundle might be realized as a particular affix.
 For systems to succeed at this type of generalization, they must treat each individual feature separately, rather than treating feature bundles as holistic. In the medium data condition for some languages, some submissions far surpassed oracle-fc. As in 2017, the most notable example of this is Basque, where oracle-fc produced a 44\% accuracy while six of the submitted systems produced an accuracy of 80\% or above. Basque is an extreme example with very large paradigms for the few verbs that inflect in the language, so the problem of generalizing correctly to unseen feature combinations is amplified.

\begin{table*}
\centering{
\begin{adjustbox}{width=0.9\textwidth}
\begin{tabular}{llll}
\toprule
& \multicolumn{3}{c}{Task 1 - Part 1} \\
& High & Medium & Low \\
\midrule
Adyghe & 100.00(uzh-2) & 94.40(uzh-1) & 90.60(ua-8)\\
Albanian & 98.90(bme-2) & 88.80(iitbhu-iiith-2) & 36.40(uzh-1)\\
Arabic & 93.70(uzh-1) & 79.40(uzh-1) & 45.20(uzh-1)\\
Armenian & 96.90(bme-2) & 92.80(uzh-1) & 64.90(uzh-1)\\
Asturian & 98.70(uzh-1) & 92.40(iitbhu-iiith-2) & 74.60(uzh-2)\\
Azeri & 100.00(axsemantics-2) & 96.00(iitbhu-iiith-2) & 65.00(iitbhu-iiith-2)\\
Bashkir & 99.90(uzh-2) & 97.30(uzh-2) & 77.80(iitbhu-iiith-1)\\
Basque & 98.90(bme-2) & 88.10(iitbhu-iiith-2) & 13.30(uzh-1)\\
Belarusian & 94.90(uzh-1) & 70.40(uzh-1) & 33.40(ua-8)\\
Bengali & 99.00(bme-3) & 99.00(uzh-2) & 72.00(uzh-2)\\
Breton & 100.00(waseda-1) & 96.00(uzh-2) & 72.00(uzh-1)\\
Bulgarian & 98.30(uzh-2) & 83.80(uzh-2) & 62.90(ua-8)\\
Catalan & 98.90(uzh-2) & 92.80(waseda-1) & 72.50(ua-8)\\
Classical-syriac & 100.00(axsemantics-1) & 100.00(axsemantics-2) & 96.00(uzh-2)\\
Cornish & --- & 70.00(uzh-1) & 40.00(ua-4)\\
Crimean-tatar & 100.00(iit-varanasi-1) & 98.00(uzh-2) & 91.00(iitbhu-iiith-2)\\
Czech & 94.70(uzh-1) & 87.20(uzh-1) & 46.50(uzh-2)\\
Danish & 95.50(uzh-1) & 80.40(uzh-1) & 87.70(ua-6)\\
Dutch & 97.90(uzh-1) & 85.70(uzh-1) & 69.30(ua-6)\\
English & 97.10(uzh-2) & 94.50(uzh-1) & 91.80(ua-8)\\
Estonian & 98.40(uzh-2) & 81.60(uzh-1) & 35.20(uzh-1)\\
Faroese & 87.10(bme-2) & 72.60(uzh-1) & 49.80(ua-8)\\
Finnish & 95.40(uzh-1) & 82.80(uzh-1) & 25.70(uzh-1)\\
French & 90.40(uzh-2) & 80.90(uzh-2) & 66.60(uzh-2)\\
Friulian & 99.00(axsemantics-2) & 97.00(iitbhu-iiith-1) & 79.00(uzh-2)\\
Galician & 99.50(uzh-1) & 90.80(uzh-1) & 61.10(uzh-2)\\
Georgian & 99.10(uzh-1) & 94.00(uzh-2) & 88.20(ua-8)\\
German & 90.20(uzh-2) & 80.10(uzh-1) & 67.10(ua-3)\\
Greek & 91.70(uzh-1) & 75.50(uzh-2) & 32.30(uzh-1)\\
Greenlandic & --- & 98.00(uzh-2) & 80.00(iitbhu-iiith-1)\\
Haida & 100.00(axsemantics-2) & 94.00(uzh-2) & 63.00(uzh-2)\\
Hebrew & 99.50(uzh-1) & 85.40(uzh-1) & 56.70(ua-8)\\
Hindi & 100.00(axsemantics-1) & 97.60(uzh-2) & 78.00(uzh-2)\\
Hungarian & 87.20(uzh-1) & 74.50(iitbhu-iiith-2) & 48.20(ua-8)\\
Icelandic & 91.30(uzh-1) & 73.80(uzh-1) & 56.20(ua-8)\\
Ingrian & --- & 92.00(uzh-2) & 46.00(iitbhu-iiith-2)\\
Irish & 91.50(uzh-2) & 77.10(uzh-1) & 37.70(uzh-1)\\
Italian & 98.00(uzh-2) & 95.10(uzh-2) & 57.40(uzh-2)\\
Kabardian & 100.00(hamburg-1) & 100.00(bme-2) & 92.00(uzh-1)\\
Kannada & 100.00(bme-3) & 94.00(uzh-2) & 61.00(uzh-1)\\
Karelian & --- & 100.00(uzh-2) & 94.00(ua-5)\\
Kashubian & --- & 88.00(bme-2) & 68.00(ua-5)\\
Kazakh & --- & 88.00(iitbhu-iiith-2) & 86.00(uzh-2)\\
Khakas & --- & 98.00(bme-3) & 86.00(iitbhu-iiith-2)\\
Khaling & 99.70(uzh-1) & 86.00(iitbhu-iiith-1) & 33.80(ua-8)\\
Kurmanji & 94.60(uzh-1) & 93.20(uzh-1) & 87.40(uzh-2)\\
Ladin & 99.00(uzh-2) & 95.00(uzh-2) & 72.00(uzh-1)\\
Latin & 78.90(bme-2) & 53.30(uzh-1) & 33.10(ua-6)\\
\bottomrule
\end{tabular}
\end{adjustbox}}
\caption{Best per-form accuracy (and corresponding system) by language. First 50 languages.}
\label{tab:bestbylanguagep1}
\end{table*}

\begin{table*}
\centering{
\begin{adjustbox}{width=0.85\textwidth}
\begin{tabular}{llll}
\toprule
& \multicolumn{3}{c}{Task 1 - Part 2} \\
& High & Medium & Low \\
\midrule
Latvian & 98.20(uzh-2) & 90.60(uzh-1) & 57.30(ua-6)\\
Lithuanian & 95.50(uzh-2) & 63.90(uzh-1) & 32.60(ua-6)\\
Livonian & 100.00(uzh-2) & 82.00(uzh-1) & 35.00(ua-8)\\
Lower-sorbian & 97.80(uzh-1) & 85.10(uzh-1) & 54.30(ua-6)\\
Macedonian & 97.40(uzh-1) & 91.60(uzh-1) & 68.80(ua-6)\\
Maltese & 97.00(uzh-2) & 95.00(uzh-1) & 49.00(ua-6)\\
Mapudungun & --- & 100.00(uzh-2) & 86.00(ua-4)\\
Middle-french & 99.30(uzh-2) & 94.50(uzh-2) & 84.50(uzh-2)\\
Middle-high-german & --- & 100.00(uzh-2) & 84.00(uzh-2)\\
Middle-low-german & --- & 100.00(iitbhu-iiith-1) & 54.00(uzh-1)\\
Murrinhpatha & --- & 96.00(uzh-2) & 38.00(ua-8)\\
Navajo & 91.00(bme-2) & 54.30(uzh-1) & 20.80(uzh-1)\\
Neapolitan & 99.00(uzh-2) & 99.00(uzh-2) & 89.00(uzh-2)\\
Norman & --- & 88.00(iitbhu-iiith-1) & 66.00(ua-4)\\
North-frisian & 96.00(bme-1) & 91.00(uzh-1) & 45.00(iitbhu-iiith-2)\\
Northern-sami & 98.30(uzh-1) & 76.10(uzh-1) & 35.80(ua-8)\\
Norwegian-bokmaal & 92.10(uzh-2) & 84.10(uzh-1) & 90.10(ua-6)\\
Norwegian-nynorsk & 94.90(uzh-2) & 67.10(uzh-1) & 83.60(ua-8)\\
Occitan & 99.00(bme-2) & 96.00(waseda-1) & 77.00(uzh-2)\\
Old-armenian & 90.40(uzh-2) & 80.20(uzh-1) & 42.00(uzh-2)\\
Old-church-slavonic & 97.00(uzh-2) & 93.00(uzh-2) & 53.00(iitbhu-iiith-2)\\
Old-english & 88.70(uzh-1) & 65.60(uzh-1) & 46.50(ua-8)\\
Old-french & 92.40(uzh-1) & 79.30(uzh-1) & 46.20(uzh-2)\\
Old-irish & --- & 40.00(uzh-1) & 8.00(baseline)\\
Old-saxon & 98.30(uzh-1) & 80.90(uzh-2) & 46.60(ua-6)\\
Pashto & 100.00(waseda-1) & 85.00(uzh-1) & 48.00(uzh-2)\\
Persian & 99.90(bme-2) & 93.40(uzh-2) & 67.60(uzh-2)\\
Polish & 93.40(uzh-2) & 82.40(uzh-2) & 49.40(ua-6)\\
Portuguese & 98.60(uzh-2) & 94.80(uzh-2) & 75.80(uzh-2)\\
Quechua & 99.90(uzh-2) & 98.90(uzh-1) & 70.20(uzh-2)\\
Romanian & 89.00(uzh-2) & 77.60(uzh-1) & 46.20(uzh-1)\\
Russian & 94.40(uzh-2) & 86.90(uzh-1) & 53.50(uzh-1)\\
Sanskrit & 96.50(uzh-1) & 85.90(uzh-2) & 58.00(uzh-1)\\
Scottish-gaelic & --- & 94.00(iitbhu-iiith-1) & 74.00(iitbhu-iiith-2)\\
Serbo-croatian & 92.40(uzh-2) & 86.10(uzh-1) & 44.80(ua-3)\\
Slovak & 97.10(uzh-1) & 78.60(uzh-1) & 51.80(uzh-2)\\
Slovene & 97.40(uzh-1) & 86.20(uzh-1) & 58.00(uzh-2)\\
Sorani & 90.60(uzh-2) & 80.20(iitbhu-iiith-2) & 40.10(uzh-1)\\
Spanish & 98.10(uzh-2) & 92.00(iitbhu-iiith-2) & 73.20(ua-8)\\
Swahili & 100.00(bme-3) & 99.00(uzh-2) & 72.00(iitbhu-iiith-2)\\
Swedish & 93.30(uzh-1) & 79.80(uzh-1) & 79.00(ua-8)\\
Tatar & 99.00(axsemantics-1) & 98.00(uzh-2) & 90.00(ua-8)\\
Telugu & --- & --- & 96.00(ua-8)\\
Tibetan & --- & 56.00(uzh-2) & 58.00(iitbhu-iiith-1)\\
Turkish & 98.50(uzh-2) & 90.70(uzh-1) & 39.50(iitbhu-iiith-2)\\
Turkmen & --- & 98.00(iitbhu-iiith-1) & 90.00(uzh-2)\\
Ukrainian & 96.20(uzh-2) & 81.40(uzh-1) & 57.10(ua-6)\\
Urdu & 100.00(iitbhu-iiith-1) & 96.80(uzh-2) & 72.50(uzh-2)\\
Uzbek & 100.00(axsemantics-1) & 100.00(axsemantics-2) & 92.00(uzh-1)\\
Venetian & 99.20(uzh-2) & 95.10(uzh-2) & 78.80(uzh-2)\\
Votic & 90.00(uzh-2) & 88.00(uzh-2) & 34.00(ua-7)\\
Welsh & 95.00(bme-3) & 85.00(bme-2) & 55.00(uzh-2)\\
West-frisian & 99.00(uzh-1) & 98.00(uzh-2) & 56.00(uzh-1)\\
Yiddish & 100.00(uzh-2) & 94.00(uzh-2) & 87.00(ua-8)\\
Zulu & 99.80(uzh-1) & 87.30(uzh-2) & 33.00(uzh-1)\\
\end{tabular}
\end{adjustbox}}
\caption{Best per-form accuracy (and corresponding system) by language. Remaining 53 languages.}
\label{tab:bestbylanguagep2}
\end{table*}

\subsection{Task 2 Results}
All systems submitted for task 2 were neural systems. All but one of the systems were encoder-decoder systems reminiscent of \newcite{kann-schutze:2016:P16-2}. The exception, \newcite{makarov-clematide:2018:K18-20},  used a neural transition-based transducer with a designated copy action, which edits the input lemma into an output form.  \autoref{tab:task2-design} details some of the design features in task 2 systems. 

{\it Predict MSD} systems predicted the MSD of the target word form based on contextual cues and used the MSD to improve performance. The system by \newcite{kementchedjhieva-bjerva-augenstein:2018:K18-20} used MSD prediction as an auxiliary task. The system by \newcite{liu-EtAl:2018:K18-20} instead converted the contextual reinflection problem into ordinary morphological reinflection. They first predicted the MSD of the target word form based on sentence context and then generated the target word form using the input lemma and the predicted MSD. 

Several systems improved upon the context model in the neural baseline system. Three systems (BME-HAS, NYU, and ZHU) used {\it subword context} models, for example, character-level models to encode context word forms, lemmata and MSDs. Many systems \cite{acs:2018:K18-20,kementchedjhieva-bjerva-augenstein:2018:K18-20,kann-lauly-cho:2018:K18-20} also used a {\it context RNN} for encoding sentence context exceeding the immediate neighboring words. \newcite{kann-lauly-cho:2018:K18-20} used {\it context attention} which refers to an attention mechanisms directed at contextual information. 

The system by \newcite{kementchedjhieva-bjerva-augenstein:2018:K18-20} was {\it multilingual} in the sense that it combined training data for all task 2 languages. 
Finally, the system by \newcite{makarov-clematide:2018:K18-20} used {\it beam search} for decoding.

Overall performance for all data settings in tracks 1 and 2 of task 2 is described in \autoref{tab:task2-overall}. For evaluation with regard to original forms, the evaluation criterion is accuracy; that is, how often a system correctly predicted the original UD form. For evaluation with regard to plausible forms, the evaluation criterion is relaxed accuracy given the set of contextually plausible forms. In other words, we measure how often the prediction was one of the variants in the set of plausible forms.

In track 1, the COPENHAGEN system is the clear winner in the high and medium data settings, whereas the UZH system is the clear winner in the low data setting. In fact, UZH is the only system which can beat the lemma copying baseline COPY-BL in the low setting. In track 2, the COPENHAGEN system and the neural baseline system NEURAL-BL deliver comparable performance in the high data setting. In the medium and low setting, the UZH system is the clear winner. Once again, the UZH system is the only system which can beat the lemma copying baseline COPY-BL in the low setting. 

\autoref{tab:task2-overall} shows that the best track 1 system  outperforms the best track 2 system for every data setting, meaning  that the additional supervision offered by context lemmata and MSDs is useful. Moreover, this effect seems to strengthen with increasing amounts of training data: the difference in performance between the best track 1 and track 2 systems for original forms in the low data setting is 3.8\%-points, in the medium setting 7.8\%-points,  and in the high setting 13.6\%-points. A further observation is that it seems to be more difficult to deliver improvements over the neural baseline system NEURAL-BL in the high setting in track 2, where NEURAL-BL in fact is one of the top two systems. This may be a result of the relatively small training sets: even in the high data setting, the training sets only contain approximately $10^5$ tokens.

The results on original and plausible forms show strong agreement. In all but one case, the same systems deliver the strongest performance for both evaluation criteria. The only exception is the Track 2 high setting where COPENHAGEN is the top system with regard to original forms and NEURAL-BL with regard to plausible forms. However, the performance of these systems is very similar. This strong agreement indicates that evaluation on plausible forms might not be necessary.

The best-performing systems for each language, track, and data setting in task 2 are given in \autoref{tab:task2-detailed}. In track 1, COPENHAGEN achieves the strongest results for most languages in the high and medium data settings, whereas UZH delivers the best performance on all languages in the low setting. In track 2, COPENHAGEN and NEURAL-BL deliver the best performance on an equal number of languages in the high setting, whereas UZH delivers best performance for most languages in the low and medium settings, and COPENHAGEN performs best for the remaining languages.

\begin{table*}
\begin{adjustbox}{width=\textwidth}
\begin{tabular}{lcccccccccccc}
\toprule
 & \multicolumn{6}{c}{Track 1} & \multicolumn{6}{c}{Track 2}\\
 & \multicolumn{3}{c}{Original} & \multicolumn{3}{c}{Plausible} & \multicolumn{3}{c}{Original} & \multicolumn{3}{c}{Plausible}\\
 & High & Medium & Low & High & Medium & Low & High & Medium & Low & High & Medium & Low \\
\cmidrule(lr{1.5pt}){1-1}\cmidrule(lr{1.5pt}){2-4}\cmidrule(lr{1.5pt}){5-7}\cmidrule(lr{1.5pt}){8-10}\cmidrule(lr{1.5pt}){11-13}
BME-HAS  & 65.69  & 45.71  & 29.34  & 73.21  & 51.28  & 32.98  & 51.83  & 36.82  & 24.71  & 60.15  & 43.80  & 31.18 \\
COPENHAGEN  & {\bf 68.51}  & {\bf 56.70}  & 24.40  & {\bf 76.10}  & {\bf 63.24}  & 26.24  & {\bf 54.93}  & 45.18  & 29.38  & 60.50  & 51.36  & 33.77 \\
CUBoulder--1  & 59.73  & 46.27  & 23.16  & 66.22  & 52.52  & 25.59  & 48.97  & 38.29  & 23.76  & 55.63  & 43.33  & 26.83 \\
CUBoulder--2  & 50.32  & 42.08  & 29.86  & 53.89  & 46.85  & 34.85  & -  & -  & -  & -  & -  & - \\
NYU  & -  & -  & -  & -  & -  & -  & -  & -  & 33.38  & -  & -  & 38.62 \\
UZH  & -  & 53.02  & {\bf 42.42}  & -  & 61.02  & {\bf 48.49}  & -  & {\bf 48.88}  & {\bf 38.60}  & -  & {\bf 55.67}  & {\bf 45.09} \\
\cmidrule(lr{1.5pt}){1-1}\cmidrule(lr{1.5pt}){2-4}\cmidrule(lr{1.5pt}){5-7}\cmidrule(lr{1.5pt}){8-10}\cmidrule(lr{1.5pt}){11-13}
NEURAL-BL  & 62.41  & 44.09  & \phantom{0}1.85  & 69.53  & 48.81  & \phantom{0}2.63  & 54.48  & 38.56  & \phantom{0}2.19  & {\bf 60.79}  & 46.74  & \phantom{0}3.11 \\
COPY-BL  & 36.62  & 36.62  & 36.62  & 42.00  & 42.00  & 42.00  & 36.62  & 36.62  & 36.62  & 42.00  & 42.00  & 42.00 \\
\bottomrule
\end{tabular}
\end{adjustbox}
\caption{Overall accuracies (in \%-points) for Tracks 1 and 2 in Task 2 for different training data settings. Results are presented separately with regard to the original forms in the UD test data sets and the manually annotated sets of plausible forms. NEURAL-BL refers to the baseline encoder-decoder system and COPY-BL to the ``lemma copying'' baseline system. Note that the output of the COPY-BL is independent of the training data and therefore results for the high, medium and low data setting are the same.}\label{tab:task2-overall}
\end{table*}

\begin{table*}
\begin{adjustbox}{width=\textwidth}
\begin{tabular}{lcccccc}
\toprule
 & \multicolumn{6}{c}{Track 1}\\
 & \multicolumn{3}{c}{Original} & \multicolumn{3}{c}{Plausible}\\
 & High & Medium & Low & High & Medium & Low\\
\cmidrule(lr{1.5pt}){1-1}\cmidrule(lr{1.5pt}){2-4}\cmidrule(lr{1.5pt}){5-7}
de  & 73.21 (BME-HAS)  & 63.90 (UZH)  & 60.06 (UZH)  & 77.55 (BME-HAS)  & 67.34 (UZH)  & 62.39 (UZH) \\
en  & 77.84 (CPH)  & 68.08 (CBL)  & 68.08 (UZH)  & 86.81 (CPH)  & 76.23 (CPH)  & 74.02 (UZH) \\
es  & 56.24 (CPH)  & 51.33 (CPH)  & 34.78 (UZH)  & 67.88 (CPH)  & 60.59 (CPH)  & 42.08 (UZH) \\
fi  & 55.27 (CPH)  & 35.71 (CPH)  & 24.90 (UZH)  & 63.02 (CPH)  & 43.07 (CPH)  & 28.97 (UZH) \\
fr  & 70.67 (CPH)  & 60.29 (CPH)  & 35.03 (UZH)  & -  & -  & - \\
ru  & 77.91 (CPH)  & 63.05 (CPH)  & 40.76 (UZH)  & 81.53 (CPH)  & 66.57 (CPH)  & 43.47 (UZH) \\
sv  & 69.26 (CPH)  & 57.66 (CPH)  & 33.30 (UZH)  & 80.32 (CPH)  & 67.23 (CPH)  & 40.00 (UZH) \\
\cmidrule(lr{1.5pt}){1-1}\cmidrule(lr{1.5pt}){2-4}\cmidrule(lr{1.5pt}){5-7}\\
 & \multicolumn{6}{c}{Track 2}\\
 & \multicolumn{3}{c}{Original} & \multicolumn{3}{c}{Plausible}\\
 & High & Medium & Low & High & Medium & Low\\
\cmidrule(lr{1.5pt}){1-1}\cmidrule(lr{1.5pt}){2-4}\cmidrule(lr{1.5pt}){5-7}
de  & 65.72 (NBL)  & 60.26 (UZH)  & 59.15 (UZH)  & 69.97 (NBL)  & 64.21 (UZH)  & 61.38 (UZH) \\
en  & 71.90 (CPH)  & 68.08 (UZH)  & 68.08 (UZH)  & 79.86 (CPH)  & 75.63 (CPH)  & 74.02 (UZH) \\
es  & 51.05 (NBL)  & 42.50 (CPH)  & 32.68 (UZH)  & 59.19 (NBL)  & 51.75 (CPH)  & 37.31 (CPH) \\
fi  & 34.82 (NBL)  & 27.06 (UZH)  & 24.40 (UZH)  & 41.17 (NBL)  & 31.89 (UZH)  & 28.21 (UZH) \\
fr  & 61.51 (CPH)  & 45.62 (CPH)  & 29.53 (CPH)  & -  & -  & - \\
ru  & 56.73 (BME-HAS)  & 54.02 (UZH)  & 28.11 (UZH)  & 60.04 (BME-HAS)  & 56.53 (UZH)  & 30.42 (UZH) \\
sv  & 55.96 (CPH)  & 47.87 (UZH)  & 32.77 (UZH)  & 66.06 (CPH)  & 56.17 (UZH)  & 39.36 (UZH) \\
\bottomrule
\end{tabular}
\end{adjustbox}
\caption{Best accuracies (in \%-points) and the for all tracks, settings and languages in task 2. The best performing system is given in parentheses. ``CPH'' refers to ``COPENHAGEN'', ``NBL'' to the neural baseline system and ``CBL'' to the ``lemma copying'' baseline system. Note, that there are no results for French with regard to plausible forms because this gold standard data set was not annotated for plausible forms (see \autoref{sec:data-task2}).}\label{tab:task2-detailed}
\end{table*}
\section{Future Directions}

In the case of inflection an interesting future topic could involve departing from orthographic representation and using more IPA-like representations, i.e.\ transductions over pronunciations.  Different languages, in particular those with idiosyncratic orthographies, may offer new challenges in this respect.\footnote{Although some recent research suggests that working with IPA or phonological distinctive features in this context yields very similar results to working with graphemes \cite{wiemerslage2018}.}

Neither task this year included unannotated monolingual corpora. Using such data is well-motivated from an L1-learning point of view, and may affect the performance of low-resource data settings, especially for the cloze task. In the inflection task, some results from last year \cite{zhou-neubig:2017:K17-20} did not see significant gains by using extra data. 

Only one team tried to learn inflection in a multilingual setting---i.e.\ to use all training data to train one model. Such transfer learning is an interesting avenue of future research, but evaluation could be difficult. Whether any cross-language transfer is actually being learned vs.\ whether having more data better biases the networks to copy strings is an evaluation step to disentangle.\footnote{This has been recently addressed by \newcite{jin2017exploring}.} 

Creating new data sets that accurately reflect learner exposure (whether L1 or L2) is also an important consideration in the design of future shared tasks.

The results for task 2 show that evaluation against the original test form versus against set of plausible forms results in a very similar ranking of systems, justifying the use of the former, much simpler, method for future shared tasks. No manual annotation would then be required for the creation of test sets, allowing the inclusion of a wider variety of languages.

In track 2 of task 2, it turned out to be difficult to achieve clear improvements over the neural baseline system. This may be a consequence of the limited amount of training data. Increasing the amount of training data is an obvious solution, but encouraging the use of external datasets for semi-supervised learning could also be an interesting direction to pursue.
Such semi-supervised methods could take the form of pretrained embeddings from monolingual corpora or more expressive models dedicated to improving morphological inflection, e.g., \citet{wolf-sonkin-EtAl:2018:Long}.

\section{Conclusion}

The CoNLL--SIGMORPHON 2018 shared task introduced a new cloze-test task with data sets for 7 languages, as well as extended the existing inflection task to include 103 languages.  In task 1 (inflection) 27 systems were submitted, while  6 systems were submitted  in task 2 (cloze test).  Neural network models prevailed in both, although significant modifications to standard architectures were required to beat a simple baseline in the low data settings in both tasks.

As in previous years, we compared inflection system performance to oracle ensembles, showing that systems possessed complementary strengths. We released the training, development, and test sets for each task, and expect these to be useful for future endeavors in morphological learning, both in sentential context and in the case of isolated word inflection.

\section*{Acknowledgements}

The first author would like to acknowledge the support of an NDSEG fellowship. MS was supported by a grant from the  Society of Swedish Literature in Finland (SLS). Several authors (CK, DY, JSG, MH) were supported in part by the Defense Advanced Research Projects Agency (DARPA) in the program Low Resource Languages for Emergent Incidents (LORELEI) under contract No.\ HR0011-15-C-0113. Any opinions, findings and conclusions or recommendations expressed in this material are those of the authors and do not necessarily reflect the views of the Defense Advanced Research Projects Agency (DARPA). NVIDIA Corp.\ donated a Titan Xp GPU used for this research.

\bibliographystyle{acl_natbib_nourl}
\bibliography{conll_results}

\clearpage
\appendix

\onecolumn
\section{Detailed Task 1 Results} \label{appends}

This section contains detailed results for each submitted system on each language. Systems are ordered by average per-form accuracy for each sub-task and data condition. Three metrics are presented for each system/language combination.
\begin{enumerate}
\item Per-Form Accuracy: Percentage of test forms inflected correctly.
\item Levenshtein Distance: Average Levenshtein distance of system-predicted form from gold inflected form.
\end{enumerate}
Scores in bold include the highest scoring non-oracle system for each language as well as any other systems that did not differ significantly in terms of per-form accuracy according to a sign test ($p >= 0.05$). Scores marked with a $\dagger$ indicate submissions that were significantly better than the feature combination oracle ($p < 0.05$), showing per-feature generalization. Scores marked with $\ddagger$ did not differ significantly from the ensemble oracle, suggesting minimal complementary information across systems.

\begin{table*}
\centering\begin{adjustbox}{width=\textwidth}
\begin{tabular}{lcccccccccccc}
\toprule
 & oracle-fc & oracle-e & uzh-01 & uzh-02 & bme-02 & iitbhu-iiith-01 & iitbhu-iiith-02 & bme-03 & bme-01 & msu-04 & iit-varanasi-01 & waseda-01\\
\midrule
Adyghe & 100.00/* & 100.00/* & \textbf{100.00/0.00}$\ddagger$ & \textbf{100.00/0.00}$\ddagger$ & 97.90/0.02 & \textbf{99.90/0.00}$\ddagger$ & \textbf{99.90/0.00}$\ddagger$ & 99.40/0.01 & 99.20/0.01 & 99.30/0.01 & \textbf{99.80/0.00}$\ddagger$ & 95.00/0.07\\
Albanian & 100.00/* & 99.60/* & 97.70/0.08 & 96.50/0.12 & \textbf{98.90/0.02} & 98.00/0.04 & 98.00/0.04 & 97.50/0.05 & 97.50/0.05 & 95.00/0.14 & 96.50/0.05 & 88.80/0.41\\
Arabic & 100.00/* & 97.40/* & \textbf{93.70/0.28} & \textbf{93.50/0.28} & 92.20/0.30 & \textbf{93.30/0.26} & \textbf{93.30/0.26} & 90.30/0.32 & 90.60/0.37 & 0.00/9.16 & 84.40/0.45 & 90.20/0.34\\
Armenian & 100.00/* & 99.60/* & \textbf{96.40/0.07} & \textbf{96.80/0.05} & \textbf{96.90/0.05} & \textbf{96.10/0.07} & \textbf{96.10/0.07} & 94.70/0.09 & 94.70/0.09 & 94.90/0.12 & 93.70/0.10 & 94.30/0.10\\
Asturian & 99.20/* & 99.80/* & \textbf{98.70/0.02} & \textbf{98.50/0.03} & \textbf{98.00/0.04} & \textbf{98.60/0.03} & \textbf{98.60/0.03} & 97.80/0.04 & 97.80/0.04 & \textbf{98.40/0.03} & \textbf{98.50/0.03} & \textbf{98.40/0.03}\\
Azeri & 100.00/* & 100.00/* & \textbf{98.00/0.02}$\ddagger$ & \textbf{98.00/0.02}$\ddagger$ & \textbf{97.00/0.05}$\ddagger$ & \textbf{99.00/0.01}$\ddagger$ & \textbf{99.00/0.01}$\ddagger$ & \textbf{98.00/0.02}$\ddagger$ & \textbf{98.00/0.02}$\ddagger$ & \textbf{98.00/0.02}$\ddagger$ & \textbf{99.00/0.02}$\ddagger$ & 93.00/0.10\\
Bashkir & 100.00/* & 100.00/* & \textbf{99.90/0.00}$\ddagger$ & \textbf{99.90/0.00}$\ddagger$ & \textbf{99.80/0.00}$\ddagger$ & \textbf{99.80/0.00}$\ddagger$ & \textbf{99.80/0.00}$\ddagger$ & \textbf{99.80/0.00}$\ddagger$ & \textbf{99.80/0.00}$\ddagger$ & \textbf{99.80/0.00}$\ddagger$ & \textbf{99.70/0.01}$\ddagger$ & 93.20/0.11\\
\bottomrule
\end{tabular}
\end{adjustbox}
\caption{Task 1 High Condition Part 1.}\label{tab:t1high1}
\end{table*}

\begin{table*}
\centering\begin{adjustbox}{width=\textwidth}
\begin{tabular}{lcccccccccccc}
\toprule
 & msu-03 & axsemantics-01 & msu-02 & racai-01 & hamburg-01 & baseline & axsemantics-02 & msu-01 & tuebingen-oslo-03 & tuebingen-oslo-02 & kucst-01 & tuebingen-oslo-01\\
\midrule
Adyghe & 96.70/0.03 & 99.00/0.01 & 99.00/0.01 & 92.00/0.08 & 93.90/0.07 & 91.60/0.08 & \textbf{99.60/0.00}$\ddagger$ & 95.40/0.05 & 97.40/0.04 & 88.20/0.16 & 98.30/0.02 & 88.90/0.24\\
Albanian & 92.70/0.17 & 88.90/0.53 & 92.40/0.20 & 95.60/0.11 & 79.90/0.54 & 79.60/0.69 & 40.90/2.66 & 83.80/0.60 & 17.90/5.12 & 10.20/5.87 & 13.80/3.68 & 11.60/5.88\\
Arabic & 0.00/9.25 & 58.20/1.38 & 0.00/4.70 & 88.50/0.40 & 74.70/0.76 & 47.10/1.49 & 34.00/2.42 & 0.00/7.79 & 79.40/0.67 & 68.70/1.05 & 0.00/10.78 & 0.10/7.55\\
Armenian & 95.40/0.08 & 90.30/0.17 & 91.30/0.15 & --- & 87.60/0.20 & 86.70/0.25 & 73.40/0.84 & 90.10/0.15 & 52.70/1.85 & 47.10/2.06 & 47.60/1.36 & 33.60/2.38\\
Asturian & \textbf{98.60/0.03} & 95.30/0.11 & 94.60/0.12 & 95.40/0.13 & 96.20/0.07 & 95.10/0.11 & 91.10/0.13 & 97.00/0.07 & 84.20/0.24 & 82.70/0.34 & 88.80/0.26 & 69.70/0.56\\
Azeri & \textbf{98.00/0.06}$\ddagger$ & 94.00/0.15 & \textbf{95.00/0.05}$\ddagger$ & 90.00/0.19 & 88.00/0.24 & 71.00/0.86 & \textbf{100.00/0.00}$\ddagger$ & 66.00/0.76 & 91.00/0.14 & 88.00/0.24 & 47.00/1.07 & 79.00/0.50\\
Bashkir & 98.90/0.03 & \textbf{99.80/0.00}$\ddagger$ & 99.10/0.02 & --- & 99.10/0.01 & 90.50/0.26 & \textbf{99.80/0.00}$\ddagger$ & 96.50/0.11 & 97.10/0.05 & 90.90/0.17 & 62.80/0.72 & 91.60/0.18\\
\bottomrule
\end{tabular}
\end{adjustbox}
\caption{Task 1 High Condition Part 2.}\label{tab:t1high2}
\end{table*}

\begin{table*}
\centering\begin{adjustbox}{width=0.98\textwidth}
\begin{tabular}{lcccccccccccc}
\toprule
 & oracle-fc & oracle-e & uzh-01 & uzh-02 & iitbhu-iiith-02 & iitbhu-iiith-01 & waseda-01 & msu-04 & msu-03 & hamburg-01 & iit-varanasi-01 & msu-02\\
\midrule
Adyghe & 100.00/* & 97.20/* & \textbf{94.40/0.06} & \textbf{94.20/0.06} & 93.40/0.07 & 93.40/0.07 & 87.80/0.16 & 80.60/0.46 & 93.50/0.07 & 92.10/0.08 & 92.90/0.08 & \textbf{94.00/0.06}\\
Albanian & 100.00/* & 95.00/* & \textbf{88.10/0.30} & \textbf{87.20/0.33} & \textbf{88.80/0.24} & 87.30/0.27 & 72.70/1.29 & 78.80/0.54 & 77.20/0.58 & 79.00/0.59 & 39.20/1.63 & 72.80/1.01\\
Arabic & 98.40/* & 86.90/* & \textbf{79.40/0.65} & \textbf{78.60/0.68} & \textbf{78.30/0.71} & 74.40/0.81 & 66.70/1.02 & 0.00/10.73 & 0.00/9.77 & 63.80/1.03 & 38.80/1.90 & 0.00/4.70\\
Armenian & 97.00/* & 97.10/* & \textbf{92.80/0.13} & \textbf{92.40/0.13} & 90.80/0.16 & 88.70/0.19 & 85.30/0.26 & 86.80/0.23 & 86.80/0.24 & 88.20/0.19 & 67.40/0.80 & 80.50/0.47\\
Asturian & 97.10/* & 96.20/* & \textbf{91.60/0.17} & \textbf{92.00/0.17} & \textbf{92.40/0.17} & \textbf{91.80/0.18} & \textbf{91.80/0.18} & 90.30/0.19 & 89.40/0.22 & \textbf{91.70/0.19} & 90.00/0.20 & 88.40/0.27\\
Azeri & 99.00/* & 99.00/* & \textbf{95.00/0.11}$\ddagger$ & \textbf{94.00/0.15}$\ddagger$ & \textbf{96.00/0.08}$\ddagger$ & \textbf{92.00/0.14} & 84.00/0.34 & \textbf{96.00/0.11}$\ddagger$ & \textbf{92.00/0.16} & 89.00/0.19 & \textbf{94.00/0.14}$\ddagger$ & 80.00/0.37\\
Bashkir & 100.00/* & 99.30/* & \textbf{96.70/0.08} & \textbf{97.30/0.06} & \textbf{96.50/0.08} & 95.00/0.11 & 87.20/0.24 & 92.00/0.22 & 92.40/0.24 & 94.10/0.13 & 95.40/0.09 & 88.30/0.25\\
\bottomrule
\end{tabular}
\end{adjustbox}
\caption{Task 1 Medium Condition Part 1.}\label{tab:t1medium1}
\end{table*}

\begin{table*}
\centering\begin{adjustbox}{width=\textwidth}
\begin{tabular}{lcccccccccccc}
\toprule
 & bme-01 & bme-03 & bme-02 & msu-01 & baseline & axsemantics-02 & axsemantics-01 & kucst-01 & tuebingen-oslo-03 & tuebingen-oslo-02 & tuebingen-oslo-01\\
\midrule
Adyghe & 90.80/0.12 & 90.50/0.13 & 93.10/0.11 & 88.90/0.13 & 84.80/0.15 & 91.90/0.08 & --- & 66.10/0.86 & 55.00/1.28 & 34.60/1.96 & 47.10/1.70\\
Albanian & 46.20/1.41 & 46.20/1.41 & 46.30/1.38 & 65.50/1.01 & 60.70/1.44 & 26.50/3.21 & --- & 10.50/3.78 & 3.30/6.25 & 3.40/6.60 & 2.00/7.59\\
Arabic & 47.00/1.63 & 47.00/1.63 & 37.20/2.76 & 0.00/7.80 & 39.50/1.83 & 22.60/3.23 & --- & 0.00/10.78 & 5.10/3.63 & 22.30/2.85 & 0.00/8.80\\
Armenian & 58.70/1.00 & 58.70/1.00 & 58.00/1.13 & 77.40/0.53 & 71.00/0.55 & 63.40/1.06 & --- & 12.30/3.18 & 9.80/3.67 & 8.30/4.11 & 8.00/4.29\\
Asturian & 87.80/0.22 & 88.60/0.22 & 86.40/0.27 & 89.60/0.21 & 89.10/0.25 & 82.50/0.35 & --- & 49.70/1.01 & 40.10/1.26 & 51.80/0.89 & 26.20/1.68\\
Azeri & 56.00/0.89 & 56.00/0.89 & 65.00/0.83 & 81.00/0.40 & 50.00/1.91 & 89.00/0.20 & --- & 11.00/2.36 & 54.00/1.10 & 46.00/1.34 & 34.00/1.74\\
Bashkir & 92.40/0.14 & 91.50/0.17 & 93.20/0.21 & 93.70/0.13 & 72.30/0.66 & 93.90/0.12 & --- & 65.40/0.84 & 59.40/1.03 & 51.40/1.17 & 45.20/1.53\\
\bottomrule
\end{tabular}
\end{adjustbox}
\caption{Task 1 Medium Condition Part 2.}\label{tab:t1medium2}
\end{table*}

\begin{table*}
\centering\begin{adjustbox}{width=0.94\textwidth}
\begin{tabular}{lcccccccccccc}
\toprule
 & oracle-fc & oracle-e & axsemantics-01 & uzh-02 & uzh-01 & ua-08 & iitbhu-iiith-02 & ua-05 & iitbhu-iiith-01 & ua-06 & ua-03 & waseda-01\\
\midrule
Adyghe & 98.30/* & 98.40/* & --- & \textbf{90.30/0.10} & \textbf{90.00/0.10} & \textbf{90.60/0.14} & \textbf{89.10/0.17} & 82.10/0.22 & 85.50/0.21 & 88.90/0.17 & 81.00/0.23 & 73.10/0.51\\
Albanian & 54.80/* & 54.90/* & --- & \textbf{35.70/1.89} & \textbf{36.40/1.87} & 20.50/4.00 & 29.70/2.00 & 21.80/4.35 & 26.40/2.20 & 20.30/4.02 & 18.90/3.03 & 24.40/4.82\\
Arabic & 54.20/* & 57.40/* & --- & \textbf{44.30/1.80} & \textbf{45.20/1.77} & 35.80/2.24 & 35.20/2.04 & 19.80/5.28 & 31.90/2.19 & 35.50/2.23 & 0.10/6.27 & 27.60/3.33\\
Armenian & 55.30/* & 76.20/* & --- & \textbf{64.60/0.82}$\dagger$ & \textbf{64.90/0.77}$\dagger$ & 49.10/1.73 & 54.40/0.92 & 50.50/1.51 & 51.60/1.02 & 43.30/1.91 & 49.90/1.53 & 37.00/2.18\\
Asturian & 65.20/* & 82.80/* & --- & \textbf{74.60/0.47}$\dagger$ & 72.50/0.52$\dagger$ & 58.00/0.95 & 70.30/0.55$\dagger$ & 67.90/0.83 & 66.30/0.62 & 56.80/1.00 & 57.00/1.05 & 58.60/1.06\\
Azeri & 71.00/* & 79.00/* & --- & \textbf{62.00/0.82} & \textbf{62.00/0.86} & 46.00/1.31 & \textbf{65.00/0.76} & 48.00/1.17 & \textbf{63.00/0.81} & 24.00/2.62 & 36.00/1.63 & 39.00/1.35\\
Bashkir & 98.00/* & 94.90/* & --- & 70.10/0.51 & 67.20/0.53 & \textbf{77.50/0.48} & \textbf{77.20/0.44} & 42.90/1.10 & \textbf{77.80/0.44} & 61.90/1.42 & 42.10/1.11 & 39.40/1.84\\
\bottomrule
\end{tabular}
\end{adjustbox}
\caption{Task 1 Low Condition Part 1.}\label{tab:t1low1}
\end{table*}

\begin{table*}
\centering\begin{adjustbox}{width=\textwidth}
\begin{tabular}{lcccccccccccc}
\toprule
 & msu-02 & ua-02 & hamburg-01 & ua-07 & baseline & ua-01 & msu-04 & msu-03 & iit-varanasi-01 & ua-04 & axsemantics-02 & tuebingen-oslo-02\\
\midrule
Adyghe & 82.50/0.23 & --- & 82.00/0.22 & 83.20/0.26 & 59.00/0.46 & 59.00/0.46 & 52.00/0.94 & 3.70/3.21 & 57.80/0.66 & 40.00/1.61 & 13.40/1.47 & 0.90/5.25\\
Albanian & 25.40/4.57 & 12.60/7.09 & 16.10/2.69 & 13.50/7.10 & 22.00/4.37 & 22.00/4.37 & 5.50/4.88 & 4.40/5.10 & 0.80/6.08 & 2.90/6.88 & 0.00/12.09 & 0.00/8.87\\
Arabic & 1.20/4.39 & 25.50/4.04 & 23.50/2.52 & 20.40/2.99 & 25.60/2.98 & 25.60/2.98 & 0.10/25.85 & 0.00/22.93 & 0.80/5.30 & 0.40/6.74 & 0.00/10.78 & 0.30/6.16\\
Armenian & 37.70/2.27 & 47.70/2.28 & 39.80/1.44 & 37.80/2.09 & 37.00/2.18 & 37.00/2.18 & 16.60/2.33 & 18.20/2.48 & 5.80/3.80 & 4.10/4.28 & 10.70/2.87 & 0.10/6.56\\
Asturian & 58.00/1.08 & 58.90/0.96 & 64.50/0.64 & 45.80/1.03 & 58.60/1.06 & 58.60/1.06 & 51.10/1.02 & 43.00/1.18 & 23.90/1.62 & 22.30/2.39 & 9.80/1.96 & 1.80/4.14\\
Azeri & 43.00/1.96 & 48.00/1.19 & 37.00/1.40 & 47.00/1.27 & 24.00/2.84 & 24.00/2.84 & 26.00/2.11 & 34.00/1.55 & 28.00/2.10 & 12.00/2.98 & 24.00/1.74 & 1.00/4.77\\
Bashkir & 45.50/1.07 & --- & 58.20/0.76 & 75.40/0.50 & 39.40/1.84 & 39.40/1.84 & 27.20/2.08 & 14.50/3.16 & 35.80/1.35 & 24.50/2.08 & 22.30/1.26 & 0.20/4.96\\
\bottomrule
\end{tabular}
\end{adjustbox}
\caption{Task 1 Low Condition Part 2.}\label{tab:t1low2}
\end{table*}

\begin{table*}
\centering\begin{adjustbox}{width=0.55\textwidth}
\begin{tabular}{lcccccccccccc}
\toprule
 & bme-01 & bme-03 & kucst-01 & bme-02 & tuebingen-oslo-03 & tuebingen-oslo-01\\
\midrule
Adyghe & 0.00/69.64 & 0.00/69.64 & 0.70/5.96 & 0.00/8.29 & 0.00/5.89 & 0.00/9.97\\
Albanian & 0.00/8.93 & 0.00/8.93 & 0.00/9.40 & 0.00/9.88 & 0.00/9.93 & 0.00/10.98\\
Arabic & 0.00/8.96 & 0.00/8.96 & 0.00/10.78 & 0.00/8.72 & 0.10/6.19 & 0.00/10.67\\
Armenian & 0.00/8.20 & 0.00/8.20 & 0.10/7.32 & 0.00/10.20 & 0.00/7.07 & 0.00/9.34\\
Asturian & 2.40/3.32 & 2.40/3.32 & 0.60/4.89 & 0.00/7.15 & 0.10/5.03 & 0.00/7.83\\
Azeri & 0.00/5.65 & 0.00/6.11 & 0.00/5.07 & 0.00/6.75 & 0.00/5.72 & 0.00/8.63\\
Bashkir & 0.00/7.63 & 0.00/7.63 & 0.50/5.17 & 0.00/7.81 & 0.00/5.58 & 0.00/7.52\\
\bottomrule
\end{tabular}
\end{adjustbox}
\caption{Task 1 Low Condition Part 3.}\label{tab:t1low3}
\end{table*}

\end{document}